\documentclass[10pt,journal,compsoc]{IEEEtran}

\ifCLASSOPTIONcompsoc
  \usepackage[nocompress]{cite}
\else
  \usepackage{cite}
\fi
\ifCLASSINFOpdf
\else
\fi
\pdfoutput=1

\usepackage{graphicx}
\usepackage[utf8]{inputenc} 
\usepackage[T1]{fontenc}    
\usepackage{url}            
\usepackage{booktabs}       
\usepackage{amsfonts}       
\usepackage{nicefrac}       
\usepackage{microtype}      
\usepackage{epsfig}
\usepackage{float}
\usepackage{multicol}
\usepackage{multirow}
\usepackage{amssymb}
\usepackage{amsmath}
\usepackage{wrapfig}
\usepackage{xspace}
\usepackage{color}
\usepackage{colortbl}
\usepackage[dvipsnames, svgnames, x11names]{xcolor}
\usepackage[misc]{ifsym}
\usepackage{makecell}
\usepackage{soul}
\usepackage{bm}
\usepackage{wrapfig}
\usepackage{graphicx}
\usepackage{bbding}
\usepackage{xcolor}
\usepackage{setspace}
\usepackage{subcaption}
\usepackage{threeparttable}

\definecolor{linkcolor}{RGB}{255,0,0}
\definecolor{urlcolor}{RGB}{255,105,180}
\definecolor{citecolor}{RGB}{66,168,235}
\usepackage[pagebackref,breaklinks=true,colorlinks,citecolor=blue,urlcolor=blue,linkcolor=blue,bookmarks=false]{hyperref}
\usepackage{adjustbox}
\usepackage{pifont}
\usepackage{caption}
\usepackage{array}
\usepackage{enumitem}
\newcolumntype{C}[1]{>{\centering\arraybackslash}p{#1}} 

\definecolor{lightgray}{rgb}{0.8, 0.8, 0.8}
\definecolor{lgray}{rgb}{0.66, 0.66, 0.66}
\definecolor{whit_tab}{RGB}{255, 255, 255}
\definecolor{gray_tab}{RGB}{235, 235, 235}
\definecolor{oran_tab}{RGB}{254, 247, 241}
\definecolor{blue_tab}{RGB}{200, 227, 245}
\definecolor{lblu_tab}{RGB}{231, 239, 248}

\usepackage[ruled,vlined,linesnumbered]{algorithm2e}

\usepackage[capitalize]{cleveref}
\crefname{section}{Sec.}{Secs.}
\Crefname{section}{Section}{Sections}
\Crefname{table}{Table}{Tables}
\crefname{table}{Tab.}{Tabs.}

\newlength\savewidth

\renewcommand{\paragraph}[1]{\vspace{1.25mm}\noindent\textbf{#1}}

\newcommand{\ie}{i.e}
\newcommand{\eg}{e.g}

\def\onedot{.\xspace}
\def\eg{\emph{e.g}\onedot} 

\def\ie{\emph{i.e}\onedot}

\def\vs{\emph{vs}\onedot}

\begin{document}
%

\title{Reference Twice: A Simple and Unified Baseline for Few-Shot Instance Segmentation}

%
%
%
%

\author{Yue Han,
        Jiangning Zhang,
        Yabiao Wang, 
        Chengjie Wang,
        Yong Liu,
        Lu Qi,
        Xiangtai Li,
        Ming-Hsuan Yang
    
\IEEEcompsocitemizethanks{
\IEEEcompsocthanksitem Y.~Han and Y.~Liu are with the Institute of Cyber-Systems and Control, Advanced Perception on Robotics and Intelligent Learning Lab (APRIL), Zhejiang University. Corresponding authors: Jiangning Zhang, Yong Liu.
\IEEEcompsocthanksitem J. Zhang, Y. Wang, and C. Wang are with Youtu Lab, Tencent.
\IEEEcompsocthanksitem X.~Li is with Tiktok, Singapore.
\IEEEcompsocthanksitem  L.Qi and M-H.Yang are with the Department of Computer Science and Engineering at University of California, Merced.
}
\thanks{The first two authors contribute equally. Xiangtai leads the project.}
}

\IEEEtitleabstractindextext{
\begin{abstract}
Few-Shot Instance Segmentation (FSIS) requires detecting and segmenting novel classes with limited support examples. Existing methods based on Region Proposal Networks (RPNs) face two issues: 1) Overfitting suppresses novel class objects; 2) Dual-branch models require complex spatial correlation strategies to prevent spatial information loss when generating class prototypes. We introduce a unified framework, \textbf{Ref}erence \textbf{T}wice (RefT), to exploit the relationship between support and query features for FSIS and related tasks. Our three main contributions are: 1) A novel transformer-based baseline that avoids overfitting, offering a new direction for FSIS; 2) Demonstrating that support object queries encode key factors after base training, allowing query features to be \textbf{enhanced twice} at both feature and query levels using simple cross-attention, thus avoiding complex spatial correlation interaction; 3) Introducing a class-enhanced base knowledge distillation loss to address the issue of DETR-like models struggling with incremental settings due to the input projection layer, enabling easy extension to incremental FSIS. Extensive experimental evaluations on the COCO dataset under three FSIS settings demonstrate that our method performs favorably against existing approaches across different shots, \eg, $+8.2/+9.4$ performance gain over state-of-the-art methods with 10/30-shots. Source code and models will be available at this  \href{https://github.com/hanyue1648/RefT}{github} site.
\end{abstract}
}

\maketitle

\IEEEdisplaynontitleabstractindextext

%
\IEEEpeerreviewmaketitle

\section{Introduction} \label{section:intro}
\begin{figure}[!h]
\centering
\captionsetup{type=figure}
\includegraphics[width=0.5\textwidth]{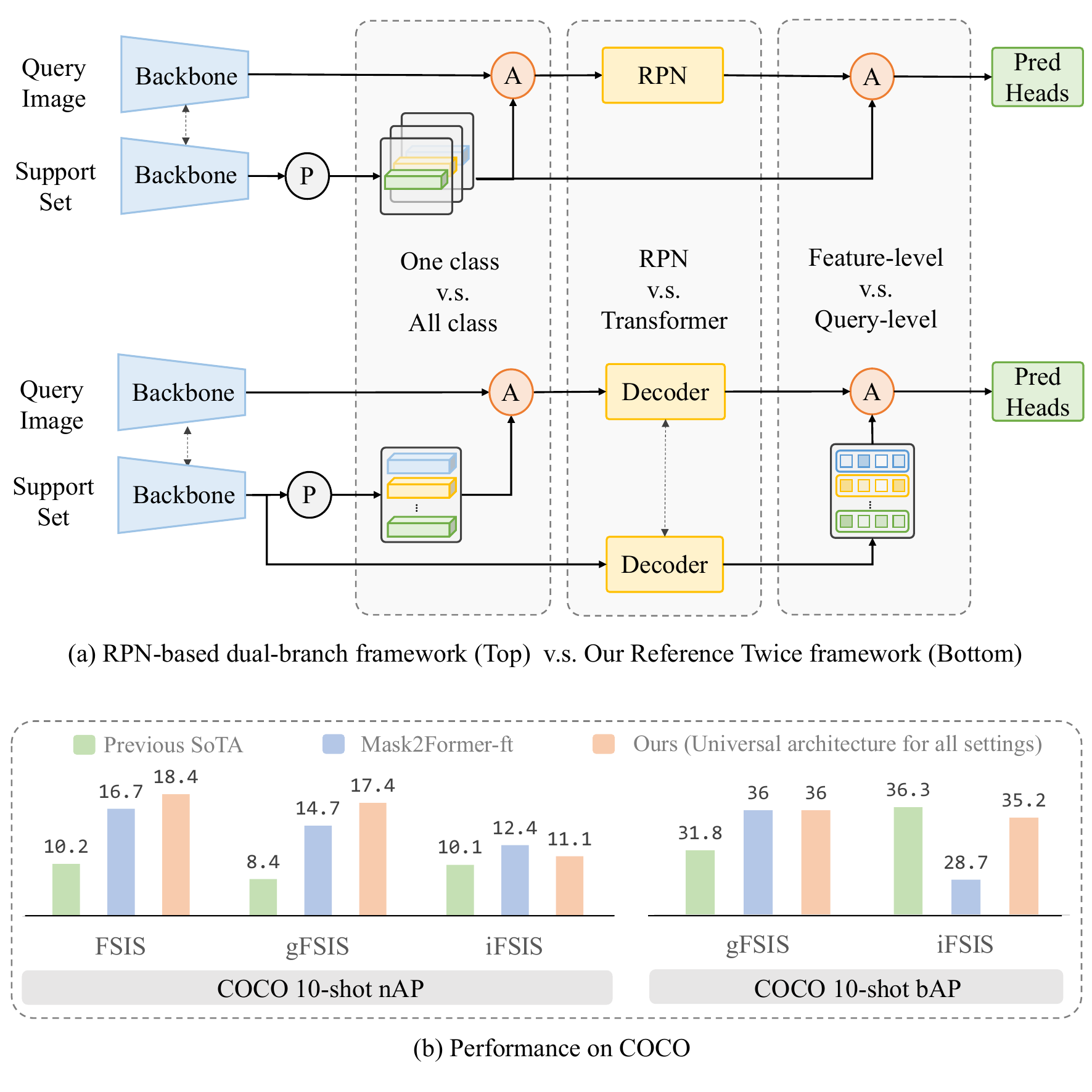} 
\captionof{figure}{(a) Existing RPN-based dual-branch framework and the proposed mask-transformer-based framework. 
Our method better utilizes the support set on feature and query levels, with only one forward pass handling all classes. 
(b) Performance on COCO 10-shot. 
Our unified baseline performs favorably in all settings. 
Here, P in the circle denotes RoI align or mask pooling, and A in the circle denotes aggregation operation.}
\label{fig:teaser}
\centering
\end{figure}  
Instance Segmentation aims to detect and segment each object in a scene, which is a core vision task widely used in scene understanding, autonomous driving, and image editing, to name a few.  
The recent years have witnessed significant success in designing models for a set of pre-defined classes for numerous vision tasks~\cite{guo2022segnext,li2023mask,jain2023oneformer,cheng2022masked,he2017mask,bolya2019yolact}. 
However, deploying these methods for real-world scenarios is challenging since they are data-hungry, and most approaches need extra mask annotations.
As such, numerous Few-Shot Learning (FSL) approaches have been developed.
With several labeled data of base classes, FSL aims at learning and predicting novel classes in the given input data (query set of images) with only a few labeled exemplars (support set of images), \ie, FSL learns a conditional model that performs prediction by referring to support images. 

To address the issue of lacking instance-wised mask annotation for novel classes, several Few-Shot Instance Segmentation (FSIS) have been  developed~\cite{yan2019meta,fan2020fgn}, mainly based on Region Proposal Networks~\cite{ren2015faster}.
%
%
Despite the promising results, these methods suffer from supervision collapse due to model overfitting. 
For example, existing RPN-based methods for FSIS tend to predict novel class foreground as the background. 
Some methods address these issues by mining additional samples~\cite{kaul2022label} and unfreezing more parameters~\cite{qiao2021defrcn}. 
While \cite{han2022few} utilizes a transformer architecture, it still relies on RPN. Existing methods \cite{zhang2022meta,fsdetr,dong2022incremental} employ the query-based detection transformer to circumvent RPN overfitting problems in FSOD. Nevertheless, they have not fully exploited the reference cues from the support branch at both feature and query levels in the dual-branch structure.
%
In this work, we develop a query-based model based on mask-transformer model, \ie, Mask2Former~\cite{cheng2022masked} to handle FSIS problems.
%
A naive solution is to freeze the model parameters after training on base classes and only fine-tune the class-specific ones on novel classes, similar to the fine-tuning based approaches using RPN-based models ~\cite{wang2020frustratingly,ganea2021incremental}. 
Although this approach performs better than RPN-based methods, we observe that misclassification occurs among many semantically correlated classes on both low level and high level, as shown in Fig.~\ref{fig:m2f_ft}. 
%

To solve the issues mentioned above, we analyze the dual-branch architecture as shown in Fig.~\ref{fig:teaser}(a), where it typically extracts the class prototypes from support images to guide the query branch detector by feature aggregation~\cite{baik2020meta,yan2019meta,fan2020fgn}. 
The positions of the feature aggregation module can be summarized before and after the RPN, and the operation methods include channel-wise multiplication, addition, subtraction, and concatenation. 
A few methods~\cite{hu2021dense, han2022meta, han2022few} show that pooling operation would lead to complete loss of spatial information when extracting class prototypes. Thus, they directly use feature maps of the two branches to capture spatial correlation via the attention mechanism.
These methods require the elaborate design of feature processing methods to avoid under-exploring visual cues from support data.
One question ensues: {\emph{How to design an effective framework better to leverage the support guidance for query-based methods for FSIS?}

In this work, we examine the properties of Mask2Former after base training and identify two critical factors that help design a solution to leverage the visual cues of support samples better.
First, we show that object queries from the support branch (with ground truth mask input) can locate objects for novel classes well, even \textit{without} the fine-tuning process on novel classes. 
We term this as \emph{support query localization}. 
Second, we calculate attention maps between object queries from support images and object queries from query images. 
We show that without fine-tuning, most object queries are highly correlated, even for most novel classes. 
We term this as \emph{support query categorization}. 
Thus, we propose query-level feature aggregation using object queries that encode high-level instance-wise categories and positioning information to complement feature-level feature aggregation using class prototypes as used in prior works~\cite{snell2017prototypical}.

Motivated by the above findings, we present a \emph{Reference Twice}} (RefT) model better to exploit the support mask information and support object queries, as shown in Fig.~\ref{fig:teaser}(b). 
RefT adopts the Meta-Learning framework~\cite{yan2019meta} with a two-stage pipeline, learning to quickly generalize to novel knowledge by referencing twice from the support branch.
For the first reference, we adopt mask pooling to crop support features to generate class prototypes in a way similar to prior works.
The difference is that we feed prototypes of all support classes simultaneously, unlike most existing methods that require multiple passes with one class at a time. 
Comparing multiple classes simultaneously helps alleviate the problem of misclassification (see Fig.~\ref{fig:m2f_ft}). 
For the second reference, as shown by \emph{support query categorization} and \emph{support query localization}, object queries from support images already encode relevant classification and localization information, and thus 
we propose a multi-head attention module to link both object queries, which enhances the classification and segmentation ability for novel classes. 
Both reference processes are well coupled and significantly improve segmenting instances from novel classes.

We note several works tackle new tasks with more stringent constraints based on FSIS, \ie, generalized FSIS (gFSIS)~\cite{fan2021generalized} and incremental FSIS (iFSIS)~\cite{ganea2021incremental} using specific designs. 
Another question arises: {\emph{Is there a simple yet unified framework for FSIS, gFSIS, and iFSIS?} 
With our framework, RefT performs well in both FSIS and gFSIS, and we introduce a class-enhanced base knowledge distillation to address the issue of adaption to the incremental settings. 
This enables our framework to easily extend to incremental FSIS(Sec.~\ref{sec:method_gen}).

%
We summarize the contributions of this work as follows:
\begin{itemize} 
%
\item  We analyze the mask-transformer-based model for FSIS and identify two key factors, support query localization, and support query categorization, which are important for guiding the transformer framework design. 
\item  We propose the RefT framework, which contains two aggregation modules at the feature and query level. 
We use cross-attention as a unified feature aggregation operation at both levels, thereby eliminating the need for complex spatial correlation interaction between features and better use of support features.
%
\item  We introduce a class-enhanced base knowledge distillation module to address model adaption issues and facilitate RefT for incremental FSIS. 
\item Extensive experimental evaluations on the COCO dataset under three FSIS settings demonstrate that our method performs favorably against existing approaches across different shots, \eg, $+8.2/+9.4$ performance gain over state-of-the-art methods with 10/30-shots. 
\end{itemize}

\section{Related Work} \label{section:related}
\noindent
\textbf{Few-Shot Classification.} A variety of techniques have been devised to extend base models for classifying new classes using limited samples.~\cite{zhang2023prompt,parnami2022learning,alayrac2022flamingo,hu2022pushing,zhang2022tip,xie2022joint,yang2022few,xu2022generating,roy2022felmi,lu2022self}. 
Existing approaches typically employ the N-way K-shot episodic training paradigm that helps adapt to multiple classification tasks. 
These approaches can be mainly based on optimization, \eg, meta learners\cite{simon2020modulating,baik2020meta,baik2020learning} and metrics, \eg, transferable embeddings~\cite{vinyals2016matching,sung2018learning,yoon2019tapnet,li2019finding}.
Various embedding methods and distance functions are explored, including extracting categorical prototypes with a fixed distance metric (cosine or Euclidean~\cite{vinyals2016matching,snell2017prototypical,sung2018learning}), utilizing task-adaptive embedding functions with a learned distance metric~\cite{yoon2019tapnet,li2019finding} and attention modules~\cite{hao2019collect,hou2019cross}. 
Instead of studying the image-level classification task, we focus on instance-level few-shot learning.

\noindent
\textbf{Few-Shot Object Detection.} A number of approaches enlarge the vocabulary of a detector with few samples~\cite{li2021few,lee2022few,zhang2022kernelized,cao2021few,li2021transformation,wu2021generalized,han2021query,chen2021dual,zhang2021accurate}. 
TFA~\cite{wang2020frustratingly} proposes a two-phase fine-tuning approach, while DeFRCN~\cite{qiao2021defrcn} decouples the training of RPN features and RoI classification. 
SRR-FSD~\cite{zhu2021semantic} combines multi-modal inputs and LVC~\cite{kaul2022label} proposes a pipeline to enlarge novel detection examples for training a more robust model. 
Meta-DETR~\cite{zhang2022meta} uses a Correlational Aggregation Module (CAM) for simultaneous aggregation between query features and support class prototypes.
While our method bears some similarties to this method, there are two main differences: (1) The support set is organized differently. 
Meta-DETR splits all classes into several class sets and requires multiple forward passes, while RefT deals with all classes at one time. 
%
(2) CAM entirely assumes the role of class matching and the sigmoid binary classification head outputs whether there is a match. 
%
This approach does not rely on embedding vectors before the classification head to capture class representative features. 
Therefore, it does not require embedding-based matching for classification and has better generalization performance, especially in low-shot scenarios.
%
However, without class representative features, the model may forget the base classes more easily. This makes it unsuitable for generalized and incremental learning settings.
To accommodate different few-shot learning settings for segmentation, RefT leverages the first reference module to filter the correct class features and employs the softmax classification head to perform class matching. 

\noindent
\textbf{Few-Shot Semantic Segmentation.} These approaches require accurate pixel-level classification of query images based only on a limited number of labeled samples~\cite{lang2023base,lang2022learning,kang2022integrative,fan2022self,liu2022intermediate,afrasiyabi2022matching,zhang2022feature,gao2022mutually}. 
To tackle this problem, prototypical feature learning  and affinity learning approaches have been developed.
Prototypical feature learning methods~\cite{tian2020prior,rakelly2018conditional,wang2019panet,zhang2020sg,liu2020part,li2021adaptive,zhang2021few} condense masked support features into single or multiple prototypes, while affinity learning schemes consider fine-grained pairwise relationships between support and query features. 
However, relying solely on prototypes can lead to information loss and performance loss, while pixel-level correlation in affinity learning methods may suffer from false matches caused by intra-class variations and cluttered backgrounds. 
Recent methods~\cite{li2021adaptive, zhang2021few} have highlighted the limitation of using a single prototype to cover all regions of an object, particularly for pixel-wise dense segmentation tasks. 
To overcome this issue, affinity learning schemes~\cite{hong2022cost,zhang2021few} mine dense correspondence between the query images and support annotations, thereby supplementing more detailed support context. AGNN~\cite{lu2021segmenting} introduces attentive graph to thoroughly examines fine-grained semantic similarities between all the possible location pairs in two data instances.

\noindent
\textbf{Few-Shot Instance Segmentation.} Existing approaches can be divided into single-branch and dual-branch architectures. 
The former~\cite{ganea2021incremental,nguyen2022ifs}  focuses on the design of the classification head, while the latter~\cite{michaelis2018one,fan2020fgn,yan2019meta} introduces an additional support branch to compute class prototypes or re-weigh vectors of support images to select target category features via feature aggregation.
For example, Meta R-CNN~\cite{yan2019meta} performs channel-wise multiplication on features belonging o region of interest, while FGN~\cite{fan2020fgn} aggregates channel-wise features at three stages, including RPN, detection head, and mask head. 
%
Several recent methods~\cite{kaul2022label, zhang2022meta, qiao2021defrcn} show that RPN-based approaches tend to mistake the novel class objects as the background. 
In contrast, RefT builds on a query-based detector to avoid this issue.

\noindent
\textbf{Generalized Few-Shot Object Detection and Instance Segmentation.}
Several approaches take knowledge contained base classes into account for generalized few-shot instance segmentation (gFSIS).
In~\cite{wang2020frustratingly} a replay strategy is used and an equal number of examples from base and new classes are used for fine-tuning to avoid class imbalance issues.
However, the problem of forgetting base classes remains unaddressed. 
Several methods tackle the catastrophic forgetting issue~\cite{fan2021generalized, ganea2021incremental} 
by weighing base classes to retain knowledge, but suffer from performance loss in generalization.  
In contrast, RefT exploit reference information from the support set to address both generalization to novel classes and forgetting of base classes. 

\noindent
\textbf{Incremental Few-Shot Object Detection and Instance Segmentation.}
Similar to the generalized setting, the incremental few-shot instance segmentation (iFSIS) also requires balancing the model performance on both base and novel classes. 
However, the incremental setting imposes more stringent constraints due to practical considerations such as data security and resource consumption. 
The learned base model does not have access to the training data, and fine-tuning can only be performed on novel class samples. 
RPN-based methods such as iMTFA~\cite{ganea2021incremental} and iFS-RCNN~\cite{nguyen2022ifs} prevent catastrophic forgetting and adapt to the incremental setting by freezing all parameters except for the classification head. 
Specifically, iMTFA employs a cosine similarity classifier and models the novel class by averaging the embedding vectors of all previous classes; and 
iFS-RCNN proposes a logit classifier based on Bayesian probabilities to discriminate between base and novel classes.

%
In recent studies, the use of transformer, \ie DETR, for incremental tasks has been explored~\cite{dong2022incremental}. However, it has been observed that designing the classification head alone is inadequate.
%
In addition to the classification head, input projection layer is also a class-specific layer, which is referred to as class adaptation layer (CAL) in this paper to emphasize its significance in novel class learning.
Without fine-tuning CAL, the model cannot learn novel classes, but fine-tuning CAL will cause serious forgetting issues. 
Incremental-DETR~\cite{dong2022incremental} addresses this by performing feature distillation on CAL, logits distillation on the classification head, and using selective search to provide pseudo labels to improve performance further. 
We show that CAL is the crucial factor in addressing the problem. 
Thus, we focus on CAL and propose a class-enhanced base class knowledge distillation module to tackle the problem.


\noindent
\textbf{Open-Vocabulary Methods.}
Open-Vocabulary learning~\cite{wu2023open, zhang2023simple, qin2023freeseg, wang2024hierarchical} increase their vocabulary size for object detection~\cite{zareian2021open, minderer2022simple, gu2021open, du2022learning, minderer2024scaling, wang2023learning, wu2023aligning, shi2023edadet, kim2023region, wang2023open, wang2023object, feng2022promptdet, yao2023detclipv2, ma2024codet, wu2023clipself} or instance segmentation~\cite{vs2023mask, wu2023betrayed} tasks by leveraging knowledge from pre-trained vision-language models. Recent efforts are dedicated to exploring knowledge distillation techniques, including the use of pre-trained CLIP~\cite{liang2023open} and pseudo-labeling~\cite{gao2022open, huynh2022open, xu2023dst}. While both open-vocabulary and few-shot settings share the same goal of accommodating novel classes, the few-shot setting specifically focuses on utilizing limited examples to learn novel classes.

\noindent
\textbf{Foundation Models.}
Numerous foundation models have been proposed for vision, language, and multi-modal tasks ~\cite{wang2023internimage,touvron2023llama,fang2023eva,zhou2023comprehensive,yu2022coca,singh2022flava, OMGSeg, li2023transformer} with 
state-of-the-art generalization capability in zero-shot scenarios. 
Recently, Segment Anything (SAM)~\cite{kirillov2023segment} develops a sophisticated data engine to collect 11 million image-mask data and trains a segmentation foundation model. 
This model introduces a novel segmentation paradigm based on prompts, including points, boxes, masks, and free-form texts, 
%
%
However, SAM does not inherently segment specific visual concepts, like precise parts in anomaly detection tasks (e.g., gears in industrial quality control)~\cite{ji2023segment}, personalized segmentation (with a specific individual)~\cite{ma2023segment,he2023accuracy,wu2023medical,zhang2023customized} and specific organs in medical imaging (\eg, tumors in MRI scans)~\cite{zhang2023personalize}.
We note that SAM and RefT address \textit{different aspects} of the segmentation problem.
RefT focuses on solving the semantic matching challenge between limited novel instance annotations and the objects to be segmented. 
In contrast, SAM is a class-agnostic segmentation model that outputs binary masks \textit{without} considering semantics. 
Thus, it cannot directly handle few-shot segmentation tasks like RefT, which benefits from its strong emphasis on semantic matching.
\section{Method} 
\label{section:method}
        


We first introduce preliminaries, including settings, baselines, and support query localization and categorization. 
Next, we introduce our RefT framework and its application for FSIS and related tasks.  
\subsection{Preliminaries}
\label{sec:pre_know}
\noindent
\textbf{Problem Settings.}\quad
For the instance segmentation tasks considered in this work, the object classes are split into   ${C}_{\textit {Base}}$ and ${C}_{\textit {Novel}}$ classes, where  ${C}_{\textit {Base}} \cap {C}_{\textit {Novel}}=\varnothing $, ${C}_{\textit {Base}} \cup {C}_{\textit {Novel }}={C}_{\textit {All}}$. 
FSIS aims to segment objects belonging to ${C}_{\textit {Test }}$ in a query image after training over abundant samples of ${C}_{\textit {Base}}$ and a few samples of ${C}_{\textit {Finetune}}$.
RefT can be applied to all three settings for FSIS in the literature. 
As shown in Tab.~\ref{tab:setting}, for FSIS, ${C}_{\textit {Finetune}}={C}_{\textit {Base}} \cup {C}_{\textit {Novel}}$, ${C}_{\textit {Test}}={C}_{\textit {Novel}}$. For gFSIS, ${C}_{\textit {Finetune}}={C}_{\textit {Base}} \cup {C}_{\textit {Novel}}$, ${C}_{\textit {Test}}={C}_{\textit {Base}} \cup {C}_{\textit {Novel}}$. 
For iFSIS, ${C}_{\textit {Finetune}}={C}_{\textit {Novel}}$, ${C}_{\textit {Test}}={C}_{\textit {Base}} \cup {C}_{\textit {Novel}}$.

\begin{table}[t]
\caption{\textbf{Different settings.} FSIS, gFSIS, iFSIS: standard, generalized, incremental few-shot instance segmentation.}
\centering
\setlength\tabcolsep{17pt}
\label{tab:setting}
\resizebox{\columnwidth}{!}{%
\begin{tabular}{@{}ccccc@{}}
\toprule   
\multirow{2}{*}{Settings} & \multicolumn{2}{c}{Fine-tune on} & \multicolumn{2}{c}{Test on}  \\ \cmidrule(l){2-3} \cmidrule(l){4-5}
                          & Base    & Novel    & Base  & Novel  \\ \midrule   

FSIS                      & \checkmark   &  \checkmark          &              & \checkmark   \\
gFSIS                     &  \checkmark    & \checkmark          & \checkmark  &  \checkmark   \\
iFSIS                     &                & \checkmark     & \checkmark   & \checkmark   \\ \bottomrule   
\end{tabular}%
}
\end{table}

\noindent
\textbf{Mask2Former FSIS Baseline.}\quad 
In this work, we use the Mask2Former~\cite{wang2020frustratingly} as our baseline. 
In the first stage, we train the model on base classes with training samples. 
In the second stage, we fix most of the parameters and only fine-tune the class-specific layers, including the input projection layer, object queries, and the class head. 
For incremental setting, we use the same settings~\cite{wang2020frustratingly} and replace the class head with a Cosine-similarity classifier. 
%
As shown in Fig.~\ref{fig:m2f_ft}, the simple fine-tuned baseline tends to misclassify semantically similar classes and confuse classes with similar shapes but completely different backgrounds. 

\begin{figure}[t]
\centering
\includegraphics[width=0.49\textwidth]{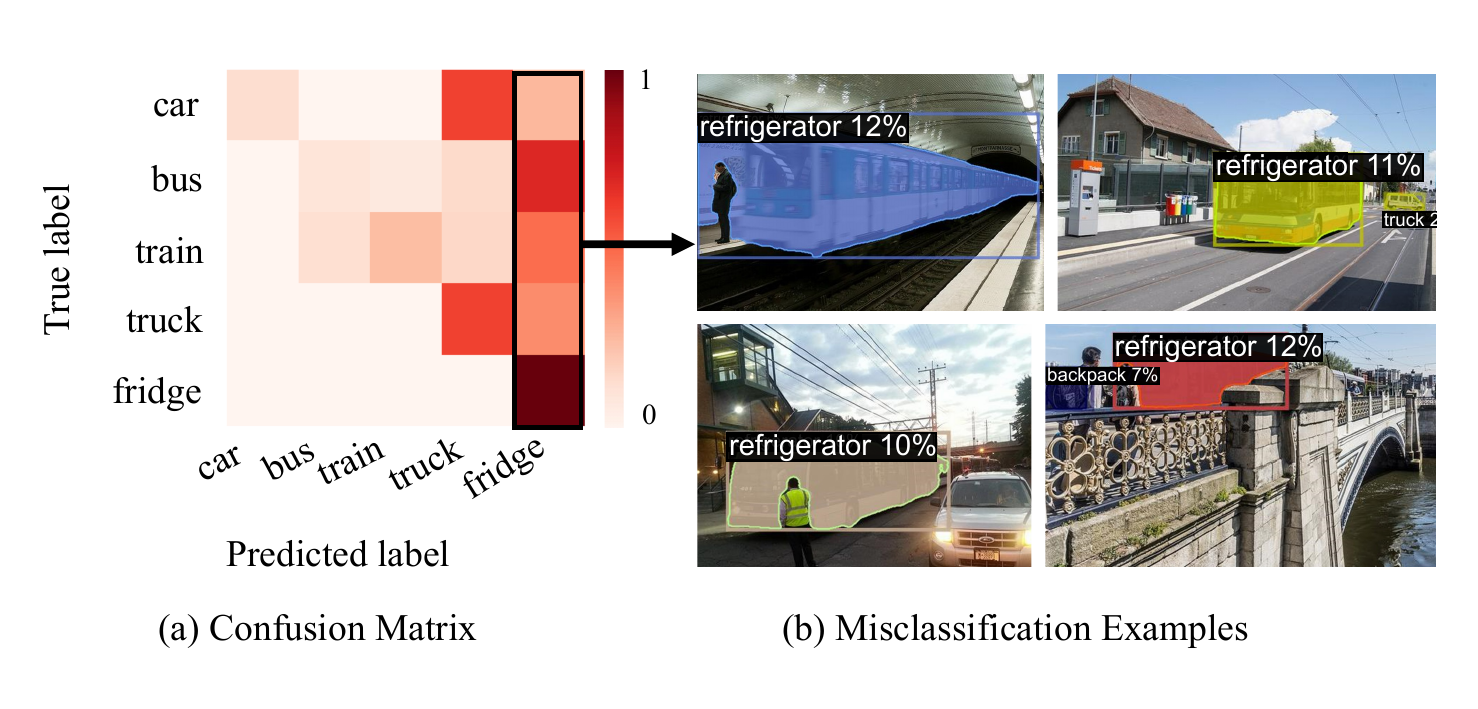}
\caption{\textbf{Baseline results.}(a) Confusion matrix and (b) visualization results of several semantically similar classes on COCO minival set (K=10) from the fine-tuned Mask2Former.} \label{fig:m2f_ft}
\centering
\end{figure}

\noindent
\textbf{Support Query Localization.}\quad 
After the base training stage, we directly infer on support images of novel classes to determine whether the model can recall novel classes with additional structural input. 
As shown in Tab.~\ref{tab:iou_table}, we find that the model is already capable of detecting and segmenting novel class objects after base training. %
This indicates that most queries have enough localization information for novel classes. 
%
Furthermore, we observe an improvement in the segmentation accuracy of more queries after the novel fine-tuning stage.

\begin{table}[t]
\centering
\caption{\textbf{Support Query Localization.} We analyze the mask quality of about 20 novel classes on COCO dataset and calculate the top-k IoU between the ground truth and predicted masks of the support branch. Object queries from support images can locate object masks for novel classes, even without novel fine-tuning. nIoU/ bIoU: average IoU of novel/ base class.}
\setlength\tabcolsep{14pt}
\resizebox{\columnwidth}{!}{%
\begin{tabular}{@{}c|cc|cc@{}}
\toprule   
\multirow{2}{*}{\# Top-k queries} & \multicolumn{2}{c|}{Base Training} & \multicolumn{2}{c}{Novel Fine-tune} \\ \cmidrule(l){2-5} 
                            & bIoU             & nIoU            & bIoU             & nIoU            \\ \midrule   
50                          & 0.54             & 0.47            & 0.59             & 0.55            \\
30                          & 0.70             & 0.64            & 0.75             & 0.71            \\
10                          & 0.84             & 0.79            & 0.84             & 0.79            \\ \bottomrule   
\end{tabular}%
}
\label{tab:iou_table}
\end{table}

\noindent
\textbf{Support Query Categorization.}\quad We show the correlation maps among the support queries in Fig.~\ref{fig:cm}, where most of these novel classes are highly correlated (more examples of correlation maps are presented in the supplementary materials). 
These correlation maps reveal \emph{support query categorization} for novel classes as the support queries have the clustering effect even without fine-tuning.
As such, these maps may be useful for query image branch training. 
In our framework, as shown on the right side of Fig.~\ref{fig:cm}, the categorization is more prominent. 
\begin{figure}[t]
\centering
\includegraphics[width=0.5\textwidth]{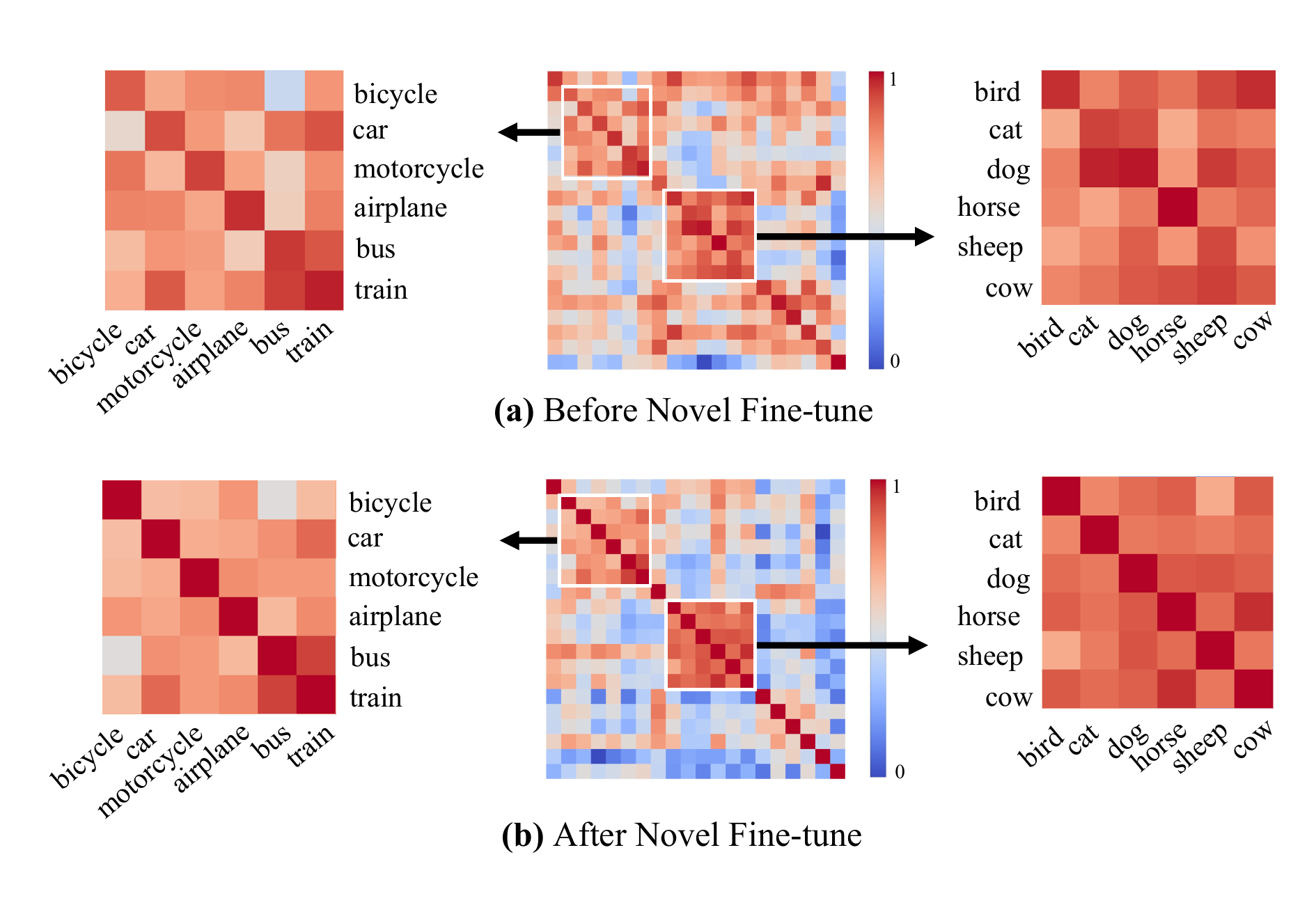}
\caption{\textbf{Support Query Categorization.} We visualize the cosine similarity of object queries of the support branch belonging to COCO 20 novel classes. 
Most object queries are roughly distinguishable, even without fine-tuning. 
We zoom in on areas that contain highly correlated and easily misclassified classes.} 
\label{fig:cm}
\end{figure}
\noindent
\subsection{Reference Twice}
\label{sec:reft}

\begin{figure*}[t]
\centering
\includegraphics[width=\textwidth]{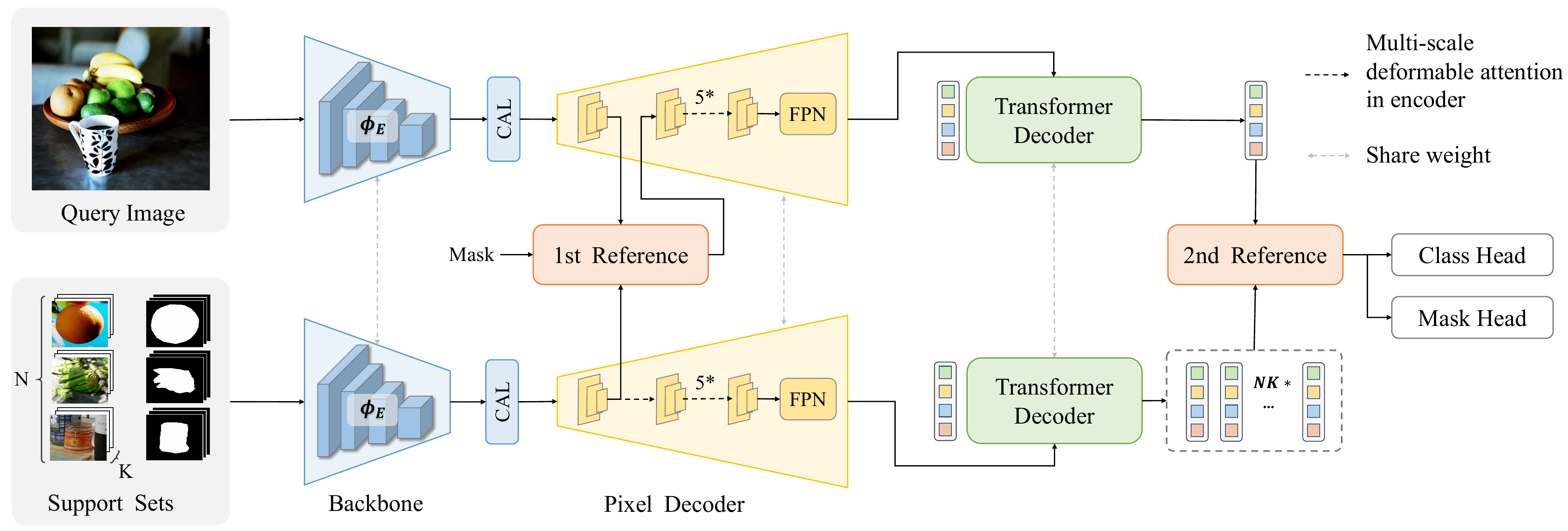}
\caption{\textbf{Architecture of the proposed \emph{Reference Twice} (RefT) for FSIS. } The query branch refers to the support branch twice on the feature and query level. 
The first reference for feature-level enhancement performs simultaneous aggregation between the query features and all adaptive class prototypes obtained through mask pooling. 
The seoncd reference module for query-level feature aggregation links object queries from the query and support branch.} 
\label{fig:method}
\end{figure*}

\begin{figure}[t]
\centering
\includegraphics[width=\columnwidth]{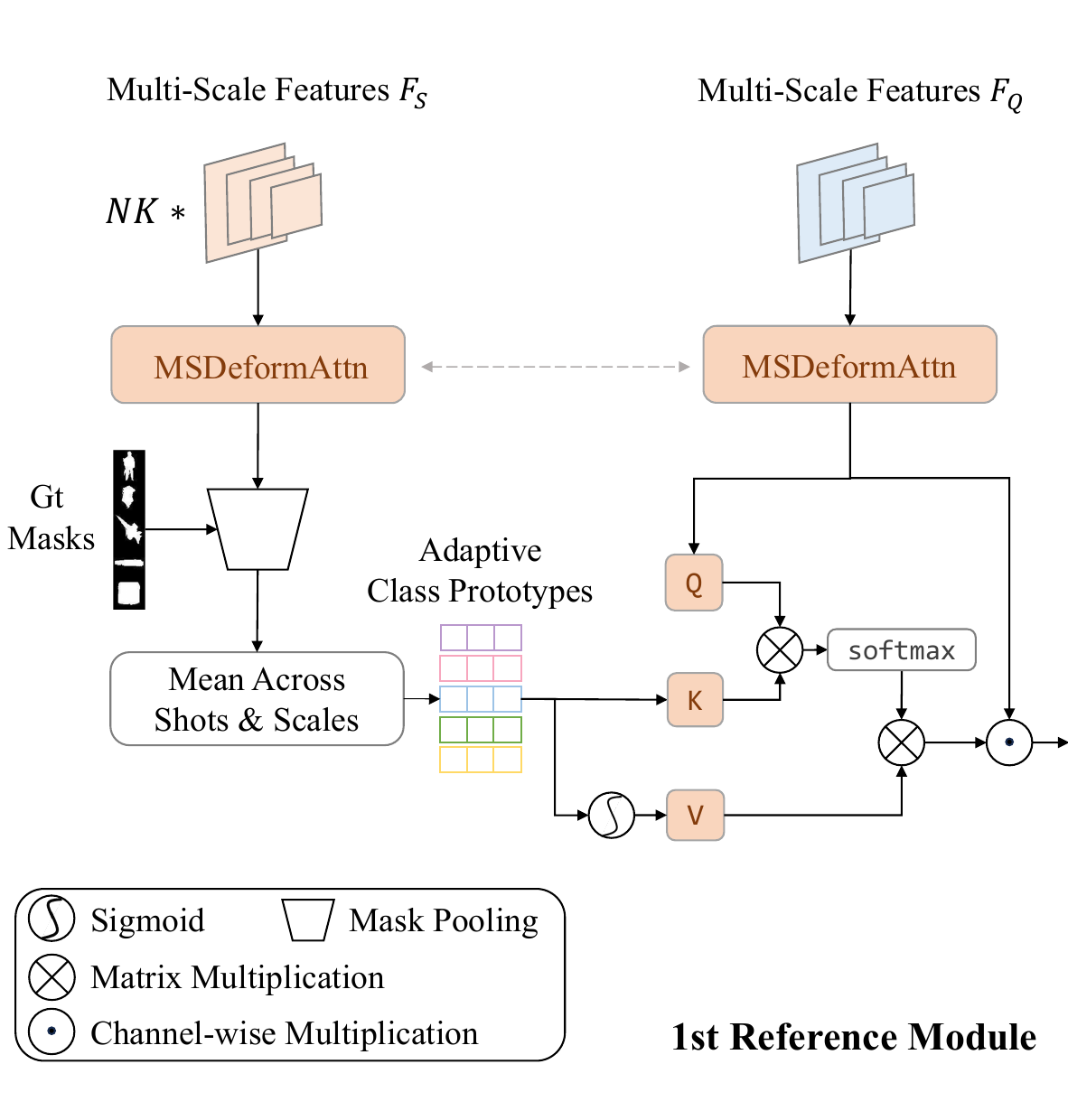}
\caption{\textbf{First Reference Module} for feature-level enhancement performs simultaneous aggregation between the query features and all adaptive class prototypes obtained through mask pooling.} \label{fig:first}
\end{figure}

\begin{figure}[t]
\centering
\includegraphics[width=\columnwidth]{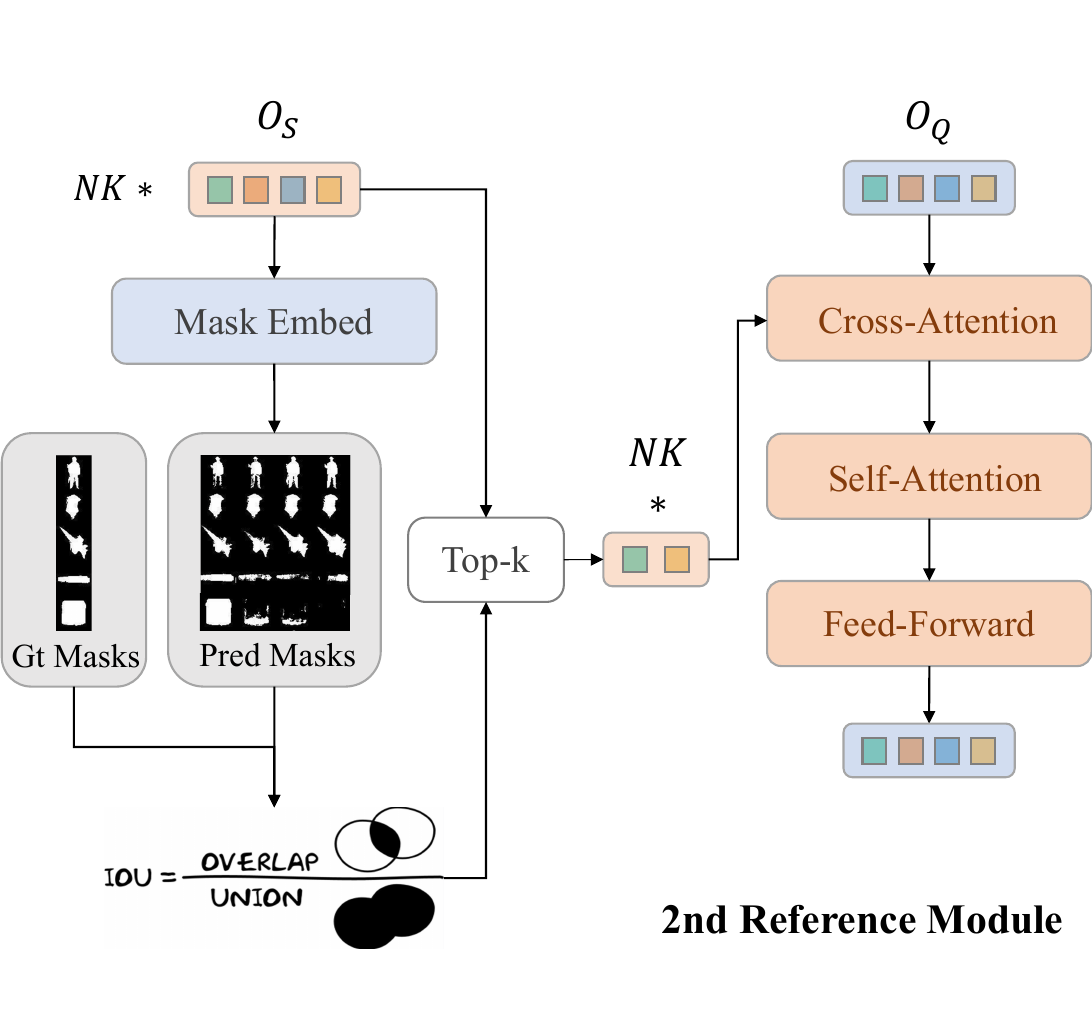}
\caption{\textbf{Second Reference Module} for query-level feature aggregation links object queries from the query and support branch.} 
\label{fig:second}
\end{figure}

\noindent
\textbf{Motivation.} As discussed above, we aim to enable the model to exploit the cues from the support queries. 
Unlike prior works~\cite{dong2022incremental} that only exploit features for segmentation, we operate on the support queries as they can encode  localization and categorization information.

\noindent
\textbf{Overview.} For illustration, we use a query image ${I}_Q$ and a support set ${S}$ with $N \times K$ examples as input, as shown in Fig.~\ref{fig:method}. 
For the support branch, we mask out the support objects and drop the background together with other objects in the image. 
As such, the support features and object queries are more discriminative and contain more accurate relevant information. 
A shared feature extractor first encodes the query and support images into the same feature space.
Subsequently, the \emph{first reference aggregation} module performs simultaneous aggregation between the query and support features. 
In this step, the query features are coarsely filtered by support categories. Then the selected query features and support features are sent to the class-agnostic pixel decoder and transformer decoder to obtain object queries from both the query and support branches.  
Next, we consider object queries of both branches for better calibration via cross-attention. 
As we mask the support images with the ground truth mask, the obtained support object queries correspond precisely to the instances of the support categories.

\noindent
\textbf{First Reference for Feature-Level Enhancement.} 
%
As shown in Fig.~\ref{fig:first}, we first use the Mask2Former encoder to extract multi-scale features.
Given the multi-scale features from the query image $\scriptstyle{F}_Q=\big\{{x}^l_Q\big\}_{l=1}^L$ and support features $\scriptstyle{F}_S^{nk}=\left\{{x}^l_S\right\}_{l=1}^L$ ($n=1, \ldots, N,k=1, \ldots, K$) (where $L$ denotes the feature level), a weight-shared multi-head deformable attention~\cite{zhu2020deformable} first encodes them into the same feature space, obtaining ${F}^{'}_Q$ and ${F}^{'}_S$. 
The features of support instances are separated from the background and other instances through mask pooling with ground truth masks on the support features of each scale, respectively. 
Then, the adaptive class prototypes for all support classes are obtained by averaging all scales per image and $K$ examples per class: 
\begin{equation}
 \centering
 {c}^n=\frac{1}{KL} \sum_{k=1}^K \sum_{l=1}^L \operatorname{MaskPool}\left({F}_S^{nk}\right), \quad n=1, \ldots, N.
\end{equation}
After that, a multi-head attention module is used to generate the reweighting matrix for aggregating ${F}_Q'$ with adaptive class prototypes ${P}=\left[{c}^1, \ldots, {c}^N\right] \in {R}^{N \times d}$:
\begin{equation}
\centering
{R}=\operatorname{softmax}\left({F}_Q' {P}^\top\right) \sigma({P}).
\end{equation}
Here, the linear projection is omitted for simplicity. The query features are then multiplied with the obtained weights along the channel dimension as below:
\begin{equation}
\centering
{F}_Q^{\textit{Enhanced}}={F}_Q' \odot {R}.
\end{equation}
Thus, the category-related query features are selected and enhanced. 
In this operation, the query branch is enhanced by support examples adaptively. We present the detailed design of choices of support features in the experiment part.

\noindent
\textbf{Second Reference for Query-Level Enhancement.} 
Given the facts in (Sec.~\ref{sec:pre_know}), we add an extra query-level enhancement module to fully leverage the relevant classification and localization information of the support set. 
We first discard support object queries with irrelevant information and only keep ones that contain instance-level category and spatial information of high quality. 
The detailed process is shown in the bottom right of Fig.~\ref{fig:method}. 
Given query object queries ${q}_Q \in \mathbb{R}^{Q \times D}$ and support
object queries ${q}_S \in \mathbb{R}^{NKQ \times D}$, $Q$ is the number of object queries per image, and $D$ is the feature dimension, we first obtain the predicted masks corresponding to the support object queries. 
Then, the top $k$ out of the $Q$ object queries per image are selected according to the IoU of the predicted and ground truth masks. 
As such, we avoid huge computation costs and also obtain better relevant cues to improve performance. 
We then use a multi-head cross-attention module to match the query and support object queries as follows:
\begin{equation}
\centering
{q}_Q^{\textit{Enhanced}} =\operatorname{softmax}\left({q}_Q \operatorname{TopK}({q}_S)^\top\right) \operatorname{TopK}({q}_S) .
\end{equation}
A multi-head self-attention module followed by one feed-forward layer is used to adapt to the following prediction heads. 
Such simple multi-head attention is good enough to link support queries to the object queries from query images, as discussed in Sec.~\ref{sec:ablation}.

\noindent
\textbf{Discussions:}
Existing methods typically rely on carefully designed spatial correlation interaction and channel aggregation to extract accurate class prototypes from features.
We do not perform query-level aggregation and keep this feature-level aggregation in the first reference module.
The reason is that object queries encode intertwined classification and localization information, unlike channel-discriminative class prototypes. 
In this work, we only need to filter the correct class features at this stage.
We use query-level aggregation in the second reference module to complement the spatial information loss in prototypes caused by pooling.
Compared with other feature-level aggregation methods, the differences are for efficiency and adaptivity.
For efficiency, we aggregate with one-forward-all-class via cross-attention, while previous dual-branch methods handle one-forward-one-class.
For adaptivity, during each iteration, the classes present in the support set change adaptively based on the classes appearing in the query image.  
We sample negative classes (classes that do not appear in the query image) and include them in the support set to mimic the scenario in the inference stage where prototypes of all classes are available including both positive and negative ones.

\begin{figure*}[t]
\centering
\includegraphics[width=\textwidth]{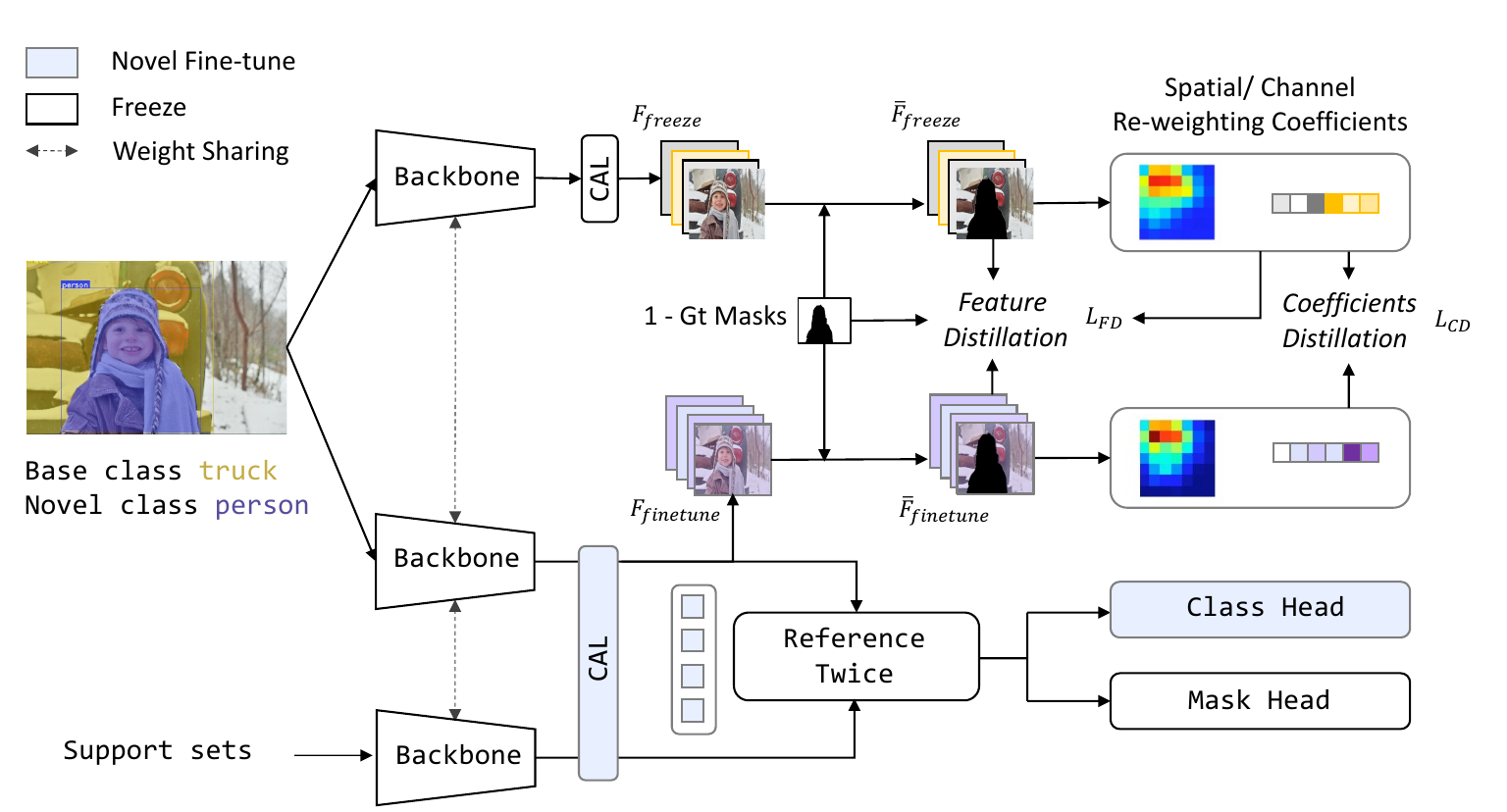}
\caption{\textbf{Adaptation to iFSIS.} In novel fine-tuning, all parameters except CAL, object queries, and the class head are frozen. The proposed Class-Enhanced BKD Loss is adopted to CAL to prevent overfitting, without hindering novel class generalization.} 
\label{fig:ifsis_arch}
\end{figure*}

\subsection{Generalization and Training Details}
\label{sec:method_gen}

\noindent
\textbf{FSIS and gFSIS.}
The common practice in prior RPN-based work is to freeze most parameters and fine-tune the class head on a balanced sampled dataset of base and novel classes to retain base class knowledge. 
However, as observed in~\cite{dong2022incremental}, the DETR-like detector can barely generalize to novel classes with the input projection layer frozen. 
We show the layer is class-specific as it transforms the channel dimension with one convolution, and channel features are essential in distinguishing different classes. 
To emphasize the significance of the layer in novel class learning, we term it as Class Adaptation Layer (CAL), as shown in Fig.~\ref{fig:method}. 
We also alleviate the catastrophic forgetting problem by simply fine-tuning CAL, object queries, and the class head with all other parameters frozen and without additional modifications. 
The same losses in Mask2Former for classification and segmentation are used. 
In addition, we classify the adaptive class prototypes in the first reference using a cosine similarity cross-entropy loss~\cite{yan2019meta,zhang2022meta} to encourage prototypes to fall into the corresponding categories.

\noindent
\textbf{iFSIS.} Existing RPN-based methods such as iMTFA~\cite{ganea2021incremental} and iFS-RCNN~\cite{nguyen2022ifs} address the catastrophic forgetting issue and adapt to the incremental setting by freezing all parameters except for the classification head. 
%
In recent studies, the use of transformer, \ie DETR, for incremental tasks has been explored~\cite{dong2022incremental}. It has been observed that fine-tuning the classification head alone is inadequate. 
The model cannot learn novel classes without fine-tuning CAL. However, fine-tuning CAL will cause serious forgetting issues. 
%
To solve this problem, Incremental-DETR~\cite{dong2022incremental} performs feature distillation on both CAL and the classification head. It further improves performance by using selective search to provide pseudo labels.
We argue that CAL is crucial for adapting DETR-like models to the incremental setting. 
As a result, to adapt RefT to iFSIS as simply as possible, we focus on the distillation on CAL without using logits distillation and selective search.
During the fine-tuning stage in iFSIS, given an input image that may contain novel class foregrounds, base class foregrounds, and backgrounds, we enforce the model to learn novel class knowledge as much as possible while retaining base class knowledge, with only access to novel class annotations. 
To this end, we use the ground truth mas to mask out novel class objects and perform feature distillation only on base class foregrounds and backgrounds:
\begin{equation}
\centering
{\Bar{F}}^{freeze}=(1-\sum_{p=1}^{P}{M}_{p}^{gt}){F}^{freeze},
\end{equation}
\begin{equation}
\centering
{\Bar{F}}^{finetune}=(1-\sum_{p=1}^{P}{M}_{p}^{gt}){F}^{finetune},
\end{equation}
where ${M}_{p}^{gt}$ is the ground truth mask of the ${p}$th foreground instance, belonging to the novel classes during fine-tuning, ${p=1,2,...,P}$. ${F}^{freeze}$, ${F}^{finetune}$ is the CAL output feature of the model frozen after base training and during fine-tuning, respectively.

%
Although without the accurate location of base class objects, we show that the learned features from the base class already indicate the approximate location. 
Thus, we directly use the relative prominence of the feature map in the spatial and channel dimensions to represent the positions of base class features and serve as reweighting weights for distillation. 
The degree of prominence is measured by the absolute mean value of the pixel or channel noarmlzed by a softmax function. 
%
The channel and spatial re-weighting coefficients ${R}_k^{C}$ and ${R}_k^{S}$ are obtained from:
\begin{equation}
\centering
{R}_k^{C} = C\cdot  \operatorname{softmax}(\frac{1}{HW}\sum_{i=1}^{H}\sum_{j=1}^{W}\left |{F}_{k, i, j}^{freeze} \right |) ,
\end{equation}
\begin{equation}
\centering
{R}_{ij}^{S} = HW\cdot  \operatorname{softmax}(\frac{1}{C}\sum_{i=1}^{C}\left | {F}_{k, i, j}^{freeze} \right |).
\end{equation}
The Re-weighted Base Class Feature Distillation Loss (FD) is formulated as:
\begin{equation}
\centering
    {\mathcal{L}}_{FD} = \sum_{k=1}^C\sum_{i=1}^H \sum_{j=1}^W{R}_k^C{R}_{i, j}^S\left({\Bar{F}}_{k, i, j}^{freeze}-{\Bar{F}}_{k, i, j}^{finetune}\right)^2,
\end{equation}
where $C$, $H$, and $W$ denote the channel, height, and width of the feature; ${F}_{k, i, j}^{freeze}$ is the masked CAL output feature of the model frozen after base training; and ${F}_{k, i, j}^{finetune}$ is the masked CAL output feature of the model during fine-tuning.

In addition, we add a Re-weighting Coefficient Distillation Loss (CD) to  enhance the base class:
\begin{align}
\centering
     {\mathcal{L}}_{CD}&=\sum_{i=1}^H \sum_{j=1}^W( {R}_{i, j}^{finetune}-{R}_{i, j}^{freeze})^{2} \\ 
    &+ \sum_{k=1}^C({R}_k^{finetune}-{R}_k^{freeze})^{2}.
\end{align}
The final \textit{Class-Enhanced Base Class Distillation Loss} (CE-BCD) is defined as:
\begin{equation}
\centering
    {\mathcal{L}}_{CE-BCD} = {\mathcal{L}}_{FD} + \lambda {\mathcal{L}}_{CD},
\end{equation}
where $\lambda$ is the hyperparameter to balance the loss terms.

\section{Experiments} \label{section:exp}

\noindent
\textbf{Dataset Setting and Metrics.} 
We evaluate the proposed method on the MS-COCO 2014 dataset~\cite{coco_dataset}with the same setups~\cite{wang2020frustratingly}, where 80 classes are divided into 2 sets, including 20 novel classes that intersect with PASCAL VOC~\cite{everingham2009pascal} and the remaining 60 base classes. 
The number of examples per class is set to $K$= \{1, 3, 5, 10, and 30\}. 
We adopt the typical metrics based on Average Precision (IoU=0.5 : 0.95), for novel and base classes, abbreviated to nAP and bAP.
As in ~\cite{ganea2021incremental}, we carry out experiments 10 times with $K$ examples of 10 seeds for each class,  
and report the averaged results. 
Since Pascal VOC has no annotations for instance segmentation, we evaluate RefT on the LVIS dataset~\cite{gupta2019lvis} to demonstrate the generalization ability. 
We use frequent ($\geq$ 100 samples per class) and common classes (11-100 samples per class) in LVIS as base classes and rare classes ($\leq$ 10 samples per class) as novel classes. 
The corresponding COCO-style metrics are abbreviated as fAP, cAP, and rAP. 
Since the number of images of rare classes is too small to make different shots, we set 
$K \leq 10$ as done in prior work~\cite{wang2020frustratingly,nguyen2022ifs}.

\begin{table*}[h!]
\caption{FSOD and FSIS results (nAP) on COCO with K = \{1, 3, 5, 10, 30\}. ``-": unavailable corresponding result. Optimal and suboptimal results are highlighted in \textbf{bold} and \underline{underline}, respectively. We use ResNet-50 as the backbone.}
\setlength\tabcolsep{14pt}
\resizebox{\textwidth}{!}{%
\begin{tabular}{@{}lcccccccccc@{}}
\toprule
& \multicolumn{5}{c}{Object Detection} & \multicolumn{5}{c}{Instance   Segmentation} \\ \cmidrule(l){2-6}\cmidrule(l){7-11}
\multirow{-2}{*}{Methods}     & 1   & 3   & 5           & 10   & 30     & 1   & 3        & 5            & 10   & 30         \\ \midrule
TFA~\cite{wang2020frustratingly}   & 1.9        & 7.0     & 9.1    & 12.1     & -            & -            & -    & -      & -    & -     \\
FSDetView~\cite{xiao2020few}  & 3.2        & 8.1         & 10.7   & 14.7    & -            & -            & -   & -    & -    & -        \\
DeFRCN~\cite{qiao2021defrcn}   & 4.8    & 10.7    & 13.6        & 16.8  & \underline{22.6}     & -            & -            & -    & -      & -       \\
FCT ~\cite{han2022few}   & 5.1          &  9.8          & 12.0                         & 15.3            &  20.2  & -            & -            & -    & - & - \\
Meta-DETR~\cite{zhang2022meta}     & \textbf{7.5} & \textbf{13.5}    & \textbf{15.4}        & \underline{19.0}  & 22.2           & - & -    & 8.1          & 10.1     & -        \\
MRCN-ft-full~\cite{he2016deep}     & 0.7 & -       & 1.3         & 2.5   &  11.1   & 0.6 & - & 1.2                     & 1.9      &  -   \\
Meta R-CNN~\cite{yan2019meta}    & - &  -       & 3.5         & 5.6  &  12.4  & - & - & 2.8                      & 4.4      &  -   \\
MTFA~\cite{ganea2021incremental}  & - & - & 6.6        & 8.5  & -        & - & -    & 6.6         & 8.4      & -   \\
iFS-RCNN~\cite{nguyen2022ifs}     & \underline{6.3} & 8.9    & 10.5       & 11.3 & 14.7     & \textbf{5.5} &7.8    & 9.4         & 10.2  & 13.1       \\ \midrule
Mask2Former-ft-full~\cite{wang2020frustratingly}   & 2.9 & 8.9 & 13.6   & 17.3    & 21.9    & 2.5 &  8.4  & 12.7   & 16.7  & 20.8  \\
Mask2Former w/ CAM   & 4.1 & 10.5 & 14.8    & 18.6    & \underline{22.9}   & 3.9 & \underline{11.1} & \underline{13.4}   & \underline{17.5} & \underline{21.6}  \\  
\cellcolor[HTML]{e9f1f6}RefT (Ours)      & \cellcolor[HTML]{e9f1f6}5.2 & \cellcolor[HTML]{e9f1f6}\underline{11.3} & \cellcolor[HTML]{e9f1f6}  \cellcolor[HTML]{e9f1f6}\underline{15.0}      &  \cellcolor[HTML]{e9f1f6}\textbf{19.3}  & \cellcolor[HTML]{e9f1f6}\textbf{24.0}        & \cellcolor[HTML]{e9f1f6}\underline{5.1}&  \cellcolor[HTML]{e9f1f6}\textbf{12.2}  &    \cellcolor[HTML]{e9f1f6}\textbf{14.2 }    & \cellcolor[HTML]{e9f1f6}\textbf{18.4}    &\cellcolor[HTML]{e9f1f6}\textbf{22.5}      \\
\bottomrule
\end{tabular}
}
\label{tab:fsis}
\end{table*}

\begin{table*}[h!]
\setlength{\abovecaptionskip}{0.2cm}
\caption{gFSOD and gFSIS results on COCO with K = \{1, 3, 5, 10, 30\}. ``-": unavailable corresponding result. Optimal and suboptimal results are highlighted in \textbf{bold} and \underline{underline}, respectively.}
\label{tab:gfsis}
\setlength\tabcolsep{10pt}
\resizebox{\textwidth}{!}{%
\begin{tabular}{@{}clcccccccccc@{}}
\toprule
                             &                            & \multicolumn{10}{c}{Object Detection}                                                                                                                                                    \\ \cmidrule(l){3-12} 
                             &                            & \multicolumn{5}{c}{nAP}                                                                    & \multicolumn{5}{c}{bAP}                                                                    \\ \cmidrule(l){3-7}\cmidrule(l){8-12}
\multirow{-3}{*}{Backbone}    & \multirow{-3}{*}{Methods} & 1            & 3                & 5                            & 10             & 30               & 1           & 3                 & 5                            & 10    & 30                       \\ \midrule
& MRCN-ft-full~\cite{he2016deep}   & 0.7          &  -            & 1.1                          & 2.3      &  -          & 17.6      &  -                    & 19.4                       & 20.6             & -        \\
& MTFA ~\cite{ganea2021incremental}   & 2.1          &  -            & 6.2                          & 8.3      &  -          & 31.7      &  -                    & 33.1                        & 34.0             & -        \\
& Retentive R-CNN ~\cite{fan2021generalized}   & -          &  -            & 8.3                          & 10.5      &  13.8         & -      &  -                    & \textbf{39.2}                        & \textbf{39.2}             & \textbf{39.3}        \\
& LVC ~\cite{kaul2022label} & - & -   & -    & \underline{17.6} & \textbf{25.5} & -  & - & -    & 29.7 & 33.3 \\\cmidrule(l){2-12}
& Mask2Former-ft-full   & 3.0 & 7.8 & 9.5    & 15.3 & 20.1 & \underline{36.7}  & \underline{36.3} & 35.9    & 36.9 & 36.2 \\  
& Mask2Former w/ CAM & \textbf{5.2} & \underline{9.4}  & \underline{11.7}    & 17.3 & 22.3 & 34.9  & 34.2 & 34.4    & 35.2 & 35.0\\ 
\multirow{-4}{*}{R-50} & \cellcolor[HTML]{e9f1f6}RefT (Ours)   & \cellcolor[HTML]{e9f1f6}\textbf{5.2}         &  \cellcolor[HTML]{e9f1f6} \textbf{10.2}               &\cellcolor[HTML]{e9f1f6} \textbf{13.1}                         & \cellcolor[HTML]{e9f1f6}\textbf{18.6}      &  \cellcolor[HTML]{e9f1f6} \underline{23.4}                 & \cellcolor[HTML]{e9f1f6}\textbf{38.4}          &    \cellcolor[HTML]{e9f1f6} \textbf{36.5}           & \cellcolor[HTML]{e9f1f6}\underline{36.0}                         &\cellcolor[HTML]{e9f1f6} \underline{37.7}       &  \cellcolor[HTML]{e9f1f6} \underline{37.2}                \\ \midrule
&  TFA ~\cite{wang2020frustratingly}     & 1.9             &  5.1        & 7.0                            & 9.1      & 12.1                  & 31.9            &  32.0          & 32.3                         & 32.4             &  34.2       \\
&   DeFRCN ~\cite{qiao2021defrcn}   & 4.8          &  \underline{10.7}           & \underline{13.6}                         & 16.8            &  21.2         & 30.4              &  32.1       & 32.6                         & 34.0    & 34.8                    \\

 &   LVC ~\cite{kaul2022label}   & -        &   -               & -                            & \underline{17.8}  &  \underline{24.5}              & -        &     -              & -                            & 31.9     &  33.0                  \\\cmidrule(l){2-12}
 &  Mask2Former-ft-full & 3.1        &  8.0             & 10.2                            & 16.5           &  20.2       & \underline{37.2}        & \underline{36.0}                & \underline{35.3}                            & \underline{36.9}    &   \underline{36.5}               \\
 & Mask2Former w/ CAM & \textbf{5.2} & 9.8 & 13.2    & 17.7 & 22.4 & 35.4  & 34.0 & 34.8    & 35.7 & 34.8\\
\multirow{-4}{*}{R-101}  & \cellcolor[HTML]{e9f1f6}RefT (Ours)   & \cellcolor[HTML]{e9f1f6}\textbf{5.2}  & \cellcolor[HTML]{e9f1f6}\textbf{10.8} & \cellcolor[HTML]{e9f1f6}\textbf{14.1}    & \cellcolor[HTML]{e9f1f6}\textbf{18.9} &  \cellcolor[HTML]{e9f1f6}\textbf{23.6}  & \cellcolor[HTML]{e9f1f6}\textbf{38.5} & \cellcolor[HTML]{e9f1f6}\textbf{36.4}  & \cellcolor[HTML]{e9f1f6}\textbf{36.2}    & \cellcolor[HTML]{e9f1f6}\textbf{37.7}  & \cellcolor[HTML]{e9f1f6}\textbf{37.4} \\ \midrule

      
                             &                            & \multicolumn{10}{c}{Instance Segmentation}                                                                                                                                                    \\ \cmidrule(l){3-12} 
                             &                            & \multicolumn{5}{c}{nAP}                                                                    & \multicolumn{5}{c}{bAP}                                                                    \\ \cmidrule(l){3-7}\cmidrule(l){8-12}
\multirow{-3}{*}{Backbone}    & \multirow{-3}{*}{Methods} & 1         & 3                  & 5                            & 10            & 30                & 1              & 3              & 5                            & 10             & 30               \\ \midrule
& MRCN-ft-full~\cite{he2016deep}   & 0.6          &  -            & 1.2                          & 1.9      &  -          & 15.6      &  -                    & 17.9                      & 18.1             & -        \\
  &MTFA~\cite{ganea2021incremental}  & 2.3           & -           & 6.4                          & 8.4     & -                    & 29.9           & -            & 31.3      & 31.8        &  -             \\ \cmidrule(l){2-12}                                           

& Mask2Former-ft-full & 2.9 & 7.6  & 8.5    & 14.7 & 17.9 & \underline{36.1} & \underline{35.6}  & \underline{33.4}   & \underline{36.0} & \underline{35.2} \\
& Mask2Former w/ CAM & \textbf{5.2}  & \underline{9.3} & \underline{11.2}    & \underline{16.1} & \underline{20.3} & 33.4  & 33.2 & 32.9    & 33.3 & 33.5\\ 
\multirow{-3}{*}{R-50}  & \cellcolor[HTML]{e9f1f6}RefT (Ours)   & \cellcolor[HTML]{e9f1f6}\textbf{5.2} & \cellcolor[HTML]{e9f1f6}\textbf{10.2} & \cellcolor[HTML]{e9f1f6}\textbf{12.4} & \cellcolor[HTML]{e9f1f6}\textbf{17.4} & \cellcolor[HTML]{e9f1f6}\textbf{21.7} & \cellcolor[HTML]{e9f1f6}\textbf{36.3} & \cellcolor[HTML]{e9f1f6}\textbf{36.5} & \cellcolor[HTML]{e9f1f6}\textbf{34.4} & \cellcolor[HTML]{e9f1f6}\textbf{36.0} & \cellcolor[HTML]{e9f1f6}\textbf{35.5} \\ \bottomrule
\end{tabular}%
}
\end{table*}
\noindent
\textbf{Implementation Details and Strong Baselines.}
We use Mask2Former~\cite{cheng2022masked} as the main module, and ResNet-50~\cite{he2016deep} is adopted as the backbone, similar to prior work for FSIS. 
For the base model, we train our model over COCO base classes for 50 epochs with a batch size of 8 on 4 RTX 3090 GPUs, using the AdamW optimizer~\cite{ADAMW} and the step learning rate schedule.
We set the initial learning rate of 0.0001 and a weight decay at the last epoch by 0.05. 
The settings remain the same in the novel fine-tuning stage.
For fair comparisons, we implement a fine-tune-based Mask2Former for FSIS similar to TFA~\cite{wang2020frustratingly}. 
We perform standard training on Mask2Former with the default settings over the COCO base classes and fine-tune the model over novel classes.

\subsection{Training and Inference Details}
\label{sec:train_inference}
 We use the episodic-training~\cite{yan2019meta} in both base training and the few-shot fine-tuning stage. 
 The training stage comprises a series of episodes $ \boldsymbol{E}_i = (\boldsymbol{I}^i_Q , \boldsymbol{S}^i)$, where $i$ indicates the $i$th episode. 
 Given a query image $\boldsymbol{I}^i_Q$, all objects present in the image belong to $\boldsymbol{N}_{pos}$ classes in $\boldsymbol{C}_{\text {train}}$. We also randomly add $\boldsymbol{N}_{neg}$ classes that are not present in the query image $\boldsymbol{I}^i_Q$. The support set $\boldsymbol{S}^i$ contains $\boldsymbol{N}$ classes, where $\boldsymbol{N} = \boldsymbol{N}_{pos} + \boldsymbol{N}_{neg}$, and varies between $\boldsymbol{N}_{pos}$ and $\boldsymbol{N}_{train}$. $K$ samples per class along with the structural annotations are provided as additional input, which makes the $N$-way $K$-shot episode $\boldsymbol{E}_i$.
In order to reduce computational costs, only the support features and queries for positive classes need to be calculated, while those for negative classes are sampled from the positive classes that are calculated in previous iterations.
 
 %
 During inference time, we compute adaptive class prototypes and object queries from support sets once and for all.
 Unlike prior works that require multiple forward passes for each query image, RefT only forwards once with all support classes, which is simpler and more efficient.

\begin{table*}[t]
\centering
\caption{iFSOD and iFSIS results on COCO with K = \{1, 3, 5, 10, 30\}. ``-": unavailable corresponding result. Optimal and suboptimal results are highlighted in \textbf{bold} and \underline{underline}, respectively. We use ResNet-50 as backbone.}
\setlength\tabcolsep{16pt}
\resizebox{\textwidth}{!}
{
\begin{tabular}{@{}lcccccccccc@{}}
\toprule
  & \multicolumn{10}{c}{Object Detection} \\ \cmidrule(l){2-11}
 & \multicolumn{5}{c}{nAP} & \multicolumn{5}{c}{bAP}    \\ \cmidrule(l){2-6} \cmidrule(l){7-11}
\multirow{-3}{*}{Methods}  & 1 & 3 & 5 & \multicolumn{1}{c}{10} & 30 & 1 & 3 & 5 & 10 & 30 \\ \midrule
Incremental-DETR~\cite{dong2022incremental} & - & - & - & \multicolumn{1}{c}{14.4} & - & - & - & - & \multicolumn{1}{c}{\underline{27.3}}  & - \\\midrule
 Mask2Former w/ Cos   & \textbf{4.9} & \textbf{8.9} & \textbf{13.7}  &  \multicolumn{1}{c}{\textbf{16.1}} & \textbf{21.2} & \underline{24.3} & \underline{24.6} & \underline{23.8} & \multicolumn{1}{c}{24.0} & \underline{24.2}\\ 
\cellcolor[HTML]{e9f1f6}RefT (Ours) w/ Cos    & \cellcolor[HTML]{e9f1f6}\underline{4.0}  & \cellcolor[HTML]{e9f1f6}\underline{8.8}  & \cellcolor[HTML]{e9f1f6}\underline{12.0} &  \multicolumn{1}{c}{\cellcolor[HTML]{e9f1f6}\underline{14.9}} & \cellcolor[HTML]{e9f1f6}\underline{18.9}  & \cellcolor[HTML]{e9f1f6}\textbf{32.2}  & \cellcolor[HTML]{e9f1f6}\textbf{32.0} 
 &  \cellcolor[HTML]{e9f1f6}\textbf{31.8} &  \multicolumn{1}{c}{\cellcolor[HTML]{e9f1f6}\textbf{33.4}}  & \cellcolor[HTML]{e9f1f6}\textbf{33.2}  \\ \midrule 
Methods                      & \multicolumn{10}{c}{Instance Segmentation}  \\ \midrule 
iMTFA~\cite{ganea2021incremental}  & 2.8 & - & 5.2 & \multicolumn{1}{c}{5.9} & - & 25.9 & - & 22.6 & \multicolumn{1}{c}{21.9} & - \\
iFS-RCNN~\cite{nguyen2022ifs}  & \textbf{4.0} & - & \underline{8.8} & \multicolumn{1}{c}{10.1} & -  & \underline{36.4} & - & \textbf{36.3} & \multicolumn{1}{c}{\textbf{36.3}} & - \\ \midrule
Mask2Former w/ Cos   & \underline{3.1} & \textbf{6.3} & \textbf{9.0} & \multicolumn{1}{c} {\textbf{12.4}} & \textbf{18.5} & 29.0 & 28.4 & 28.4 & \multicolumn{1}{c}{28.7} & 28.6 \\ 
\cellcolor[HTML]{e9f1f6}RefT (Ours) w/ Cos    & \cellcolor[HTML]{e9f1f6}\underline{3.1} & \cellcolor[HTML]{e9f1f6}\underline{6.0} & \cellcolor[HTML]{e9f1f6}\underline{8.8} &  \multicolumn{1}{c}{\cellcolor[HTML]{e9f1f6}\underline{11.1}} & \cellcolor[HTML]{e9f1f6}\underline{17.7} & \cellcolor[HTML]{e9f1f6}\textbf{37.0} & \cellcolor[HTML]{e9f1f6}\underline{36.3} & \cellcolor[HTML]{e9f1f6}\underline{35.3} &  \multicolumn{1}{c}{\cellcolor[HTML]{e9f1f6}\underline{35.2}}  & \cellcolor[HTML]{e9f1f6}\underline{32.1}  \\\bottomrule 
\end{tabular}
}
\label{tab:ifsis}
\end{table*}

\begin{table}[t]
\centering
\caption{Comparison of different feature aggregation operations for the second reference in gFSIS on COCO with K = 10. We use ResNet-50 as the backbone.}
\setlength\tabcolsep{34pt}
\resizebox{\columnwidth}{!}{%
\begin{tabular}{@{}lcc@{}}
\toprule 
\multirow{2}{*}{Methods} & \multicolumn{2}{c}{Instance Segmentation} \\ \cmidrule(l){2-3} 
                         & nAP                 & bAP                 \\ \midrule 
RefT                     & 17.4               & 36.0                 \\
w/ Adaptive Conv          & 16.8               & 35.4                \\
w/ Auxiliary Loss        & 17.5               & 35.2               \\ \bottomrule 
\end{tabular}%
}
\label{tab:operation}
\end{table}

\subsection{Main Results on COCO}
\noindent
\textbf{FSIS Results.}
Tab.~\ref{tab:fsis} shows that our method significantly outperforms previous works based on Mask R-CNN, which is expected given a more robust base model. 
For a fair comparison, we examine the Mask2Former baseline, and RefT still demonstrates noticeable performance gains. 
We also compare with the DETR-based approach, Meta-DETR~\cite{zhang2022meta}, which achieves state-of-the-art results in FSOD. 
Our method yields substantially better results in FSIS and comparable or slightly improved results in FSOD.

We incorporate the Correlational Aggregation Module (CAM) from Meta-DETR into Mask2Former to further align the base model. Our first reference module is similar to CAM, as both use the attention mechanism for aggregation. 
However, CAM introduces a set of task encodings for each support class and maps them to instances, acting as a classifier. Following previous works, we maintain the classical design choice in the first reference and still outperform the stronger baseline. This highlights the superiority of our query-level aggregation module in supporting the feature-level one.

\noindent
\textbf{gFSIS Results.} Few studies address the more challenging gFSIS, which necessitates retaining base class knowledge. 
Tab.~\ref{tab:gfsis} demonstrates that RefT consistently surpasses recent state-of-the-art methods in both gFSOD and gFSIS. 
RefT significantly outperforms Mask2Former with CAM on base classes, highlighting the effectiveness of our feature- and query-level modules.
We further explore alternative designs for our second reference module in Tab.~\ref{tab:operation}. First, we replace cross-attention with adaptive convolution. 
Next, we introduce an auxiliary loss, as in ~\cite{sung2018learning}, to enable the classification of concatenated queries and support branch object queries.
Despite these modifications, no improvements are observed. This suggests that the simple and effective aggregation design is adequate for utilizing high-level information to support branch object queries.

\noindent
\textbf{iFSIS Results.}
In Tab.~\ref{tab:ifsis}, we first adapt Mask2Former to the iFSIS setting as our baseline by replacing the fully connected classifier with a cosine similarity classifier, as in iMTFA~\cite{ganea2021incremental}, and fine-tuning both CAL and the class head while keeping all other parameters frozen after base training. 
A challenge arises as the class-specific CAL needs fine-tuning to enable novel class learning. 
However, without access to base class samples, CAL may overfit and experience catastrophic forgetting. 
Introducing our Class Enhanced Base Knowledge Distillation Loss addresses this issue and achieves results comparable to recent state-of-the-art methods. 
Incremental-DETR~\cite{dong2022incremental} proposes a similar loss, but our approach is simpler and equally effective.

\begin{table}[t]
\caption{FSIS, gFSIS and iFSIS results on LVIS with K$\leq$10. ``-": unavailable corresponding result. Optimal and suboptimal results are highlighted in \textbf{bold} and \underline{underline}.}
\label{tab:lvis}
\resizebox{\columnwidth}{!}{%
\begin{tabular}{@{}lccccccc@{}}
\toprule 
Settings            & FSIS  & \multicolumn{3}{c}{gFSIS} & \multicolumn{3}{c}{iFSIS} \\ \cmidrule(l){1-1}\cmidrule(l){2-2}\cmidrule(l){3-5}\cmidrule(l){6-8} 
Test on             & rAP   & rAP    & cAP     & fAP    & rAP    & cAP     & fAP    \\ \midrule 
MRCN-ft-full~\cite{he2016deep} & 18.3 & 12.8  & 25.4   & 27.8  & -      & -       & -      \\
iFS-RCNN~\cite{nguyen2022ifs}  & 21.1 & -      & -       & -      & 18.3  & \underline{26.3}   & \textbf{28.5}  \\ \midrule

M2Former-ft-full    & \underline{22.0} & \underline{21.2}  & \underline{26.9}   & \underline{28.3}  & \textbf{19.3}      & 19.9       & 22.00      \\
\cellcolor[HTML]{e9f1f6}Reft (Ours)         & \cellcolor[HTML]{e9f1f6}\textbf{23.5} & \cellcolor[HTML]{e9f1f6}\textbf{22.1}  & \cellcolor[HTML]{e9f1f6}\textbf{27.1}   & \cellcolor[HTML]{e9f1f6}\textbf{28.6}  & \cellcolor[HTML]{e9f1f6}\underline{18.6}       & \cellcolor[HTML]{e9f1f6}\textbf{26.9}       & \cellcolor[HTML]{e9f1f6} \underline{28.1}      \\ \bottomrule 
\end{tabular}%
}
\end{table}

\begin{table}[t]
\centering
\caption{Cross-domain evaluation on COCO2FSS-1000 with K = \{1, 3, 5\}.  Optimal and suboptimal results are highlighted in \textbf{bold} and \underline{underline}, respectively. We use ResNet-50 as the backbone.}
\setlength\tabcolsep{21pt}
\resizebox{\columnwidth}{!}{%
\begin{tabular}{@{}lccc@{}}
\toprule
 & \multicolumn{3}{c}{Instance   Segmentation} \\ \cmidrule(l){2-4} 
\multirow{-2}{*}{Methods}           & 1   & 3        & 5               \\ \midrule

MRCN-ft-full~\cite{he2016deep} & 79.7 & 80.3  & 81.1     \\
iFS-RCNN~\cite{nguyen2022ifs}  & 81.5 & 82.2     & 83.6      \\ \midrule
Mask2Former-ft-full~\cite{wang2020frustratingly}    & \underline{81.9} & \underline{82.5} & \underline{83.4}    \\  
\cellcolor[HTML]{e9f1f6}RefT (Ours)     & \cellcolor[HTML]{e9f1f6}\textbf{82.7}    &\cellcolor[HTML]{e9f1f6}\textbf{83.3}  & \cellcolor[HTML]{e9f1f6}\textbf{84.1}    \\
\bottomrule
\end{tabular}
}
\label{tab:fss1000}
\end{table}

\begin{table}[t]
\centering
\caption{Ablation study on each component.}
\setlength\tabcolsep{5pt}
\resizebox{\columnwidth}{!}{%
        \begin{tabular}{@{}ccccc@{}}
        \toprule 
        Baseline            & +first reference  & +second reference   & nAP   & bAP   \\ \midrule 
        Mask2Former-ft              &           &            & 14.7  & 36.0  \\
                            &\checkmark &            & 15.6  & 36.3  \\
                            &           & \checkmark & 16.3  & 35.5  \\
                            &\checkmark & \checkmark & 17.4  & 36.0  \\ \bottomrule 
        \end{tabular}%
}
\label{tab:abl_component}
\end{table}

\begin{table}[t]
\centering
\caption{Support features in first reference}
\setlength\tabcolsep{52pt}
\resizebox{\columnwidth}{!}{%
            \begin{tabular}{@{}ccc@{}}
            \toprule 
            Method            & nAP                  & bAP                 \\ \midrule 
            Res3              & 17.2                 & 36.6                \\
            Res4              & 17.0                 & 36.0                \\
            Res5              & 16.4                 & 35.2                \\
            Enc1              & 17.4                 & 36.0               \\ \bottomrule 
            \end{tabular}%
}
\label{tab:abl_feat_sel}
\end{table}

\begin{table}[t]
\centering
\caption{Support features in second reference}
\setlength\tabcolsep{42pt}
\resizebox{\columnwidth}{!}{%
        \begin{tabular}{@{}ccc@{}}
        \toprule 
        Features    & nAP   & bAP                        \\ \midrule 
        Feature-level & 16.3 & 35.7 \\
        Query-level & 17.4 & 36.0                      \\ \bottomrule 
        \end{tabular}%
}
\label{tab:abl_query_or_img}
\end{table}

\begin{table}[t]
\centering
\caption{Effect of fine-tuning CAL.}
\setlength\tabcolsep{45pt}
\resizebox{\columnwidth}{!}{%
        \begin{tabular}{@{}ccc@{}}
        \toprule 
        Fine-tune    & nAP  & bAP \\ \midrule 
         & 2.6  & 39.4  \\
        \checkmark   & 17.4 & 36.0 \\ \bottomrule 
        \end{tabular}%
}
\label{tab:abl_cal}
\end{table}

\begin{table}[t]
\centering
\caption{Pooling in first reference}
\setlength\tabcolsep{45pt}
\resizebox{\columnwidth}{!}{%
            \begin{tabular}{@{}ccc@{}}
            \toprule 
            Method    & nAP & bAP \\ \midrule 
            GAP & 17.0    & 35.4    \\
            RoiAlign & 17.2    &  35.3   \\
            MaskPool                   & 17.4  & 36.0  \\ \bottomrule 
            \end{tabular}%
}
\label{tab:abl_pooling}
\end{table}

\begin{table}[t]
\centering
\caption{Query selection in second reference}
\setlength\tabcolsep{48pt}
\resizebox{\columnwidth}{!}{%
    \begin{tabular}{@{}ccc@{}}
    \toprule 
    Sort by   & nAP   & bAP                      \\ \midrule 
    Score     & 17.2  & 35.8                    \\
    Mask      & 17.4  & 36.0                      \\ \bottomrule 
    \end{tabular}%
}
\label{tab:abl_query_sel}
\end{table}

\begin{table}[t]
\centering
\caption{Number of object queries in the second reference}
\setlength\tabcolsep{20pt}
\resizebox{\columnwidth}{!}{%
\begin{tabular}{@{}ccccc@{}}
\toprule 
\# Queries & 3   &10       &50  &100\\ \midrule 
nAP               & 17.4    &17.4    &17.5  & 16.8    \\
bAP               & 35.8    &36.0    &35.8  & 35.5  \\

\bottomrule 
\end{tabular}%
}
\label{tab:abl_query_num}
\end{table}

\begin{table}[t]
\caption{\    {Ablation study on first reference \textit{vs.} CAM} }
\setlength\tabcolsep{8pt}
\label{tab:cam}
\resizebox{\columnwidth}{!}{%
\begin{tabular}{@{}cccl@{}}
\toprule 
\multirow{2}{*}{Methods} & Support Set \& Loss & \multirow{2}{*}{nAP}     & \multirow{2}{*}{bAP}     \\
                        & Aligned with        &                          &                          \\ \midrule 
Meta-DETR~\cite{zhang2022meta} & Meta-DETR           & 10.1                     & \multicolumn{1}{c}{-}    \\
RefT w/ CAM             & Meta-DETR           & \multicolumn{1}{l}{16.3} & 26.8                     \\
RefT w/ CAM             & RefT                & 15.2                     & \multicolumn{1}{c}{32.3} \\
RefT w/ first reference        & RefT                & 17.4                     & \multicolumn{1}{c}{36.0} \\ \bottomrule 
\end{tabular}%
}
\end{table}

\begin{table}[t!]
\caption{\    {Effect of fine-tuning different layers.}}
\label{tab:finetune}
\setlength\tabcolsep{3pt}
\resizebox{\columnwidth}{!}{%
\begin{tabular}{@{}ccccccc@{}}
\toprule 
\multicolumn{1}{l}{\multirow{2}{*}{Res5}} & \multicolumn{1}{l}{\multirow{2}{*}{CAL}} & Pixel                     & Transformer               & \multicolumn{1}{l}{Object Queries} & \multicolumn{1}{l}{\multirow{2}{*}{nAP}} & \multicolumn{1}{l}{\multirow{2}{*}{bAP}} \\
\multicolumn{1}{l}{}                      & \multicolumn{1}{l}{}                     & Decoder                   & Decoder                   & \multicolumn{1}{l}{\& Class Head}                            & \multicolumn{1}{l}{}                     & \multicolumn{1}{l}{}                     \\ \midrule 
                                          &                                          & \checkmark &                           & \checkmark                        &1.8                  &31.0                                      \\
                                          &                                          & \checkmark & \checkmark & \checkmark                        &2.5                                     &17.2                                      \\
\checkmark                 &                                          &                           &                           & \checkmark                        &11.7             &32.9                                   \\
                                          & \checkmark                &                           &                           & \checkmark                        &17.4                                    &36.0                                      \\ \bottomrule 
\end{tabular}%
}
\end{table}

\begin{table}[t]
\caption{Ablation of the Class-Enhanced BKD Loss}
\centering
\setlength\tabcolsep{20pt}
\label{tab:abl_ifsis}
\resizebox{\columnwidth}{!}{%
\begin{tabular}{@{}lcc@{}}
\toprule 
Methods        & nAP & bAP \\ \midrule 
RefT & 12.4  & 28.7\\
RefT w/ Weight Regularization & 11.5    & 29.4 \\
RefT w/ BKD Loss & 10.1 & 33.2 \\
RefT w/ Class-Enhanced BKD Loss & 11.1  & 35.2 \\ \bottomrule 
\end{tabular}%
}
\end{table}

\begin{table}[t]
\caption{Employing different class heads for iFSIS.}
\centering
\setlength\tabcolsep{18pt}
\label{tab:classhead}
\resizebox{\columnwidth}{!}{%
\begin{tabular}{@{}lcc@{}}
\toprule 
Methods     & nAP & bAP \\ \midrule 
RefT w/ Cosine Similarity Classifier      & 11.1 & 35.2 \\
RefT w/ Logit Classifier    & 12.7  & 36.0 \\\bottomrule 
\end{tabular}%
}
\end{table}

\subsection{Results on LVIS}
\noindent
Tab.~\ref{tab:lvis} reports our results on LVIS with K$\leq$10 on FSIS, gFSIS, and iFSIS. 
Compared with the Mask2Former baseline, RefT improves by 1.5 and 0.9 on FSIS and gFSIS, respectively. In the more challenging iFSIS, RefT effectively retains base class knowledge, achieving a substantial increase of 7.0 cAP and 5.9 fAP, while maintaining a comparable performance on novel classes.

\subsection{Cross-domain Evaluation on COCO2FSS-1000}
\noindent
To demonstrate the scalability and generalization ability of RefT to new datasets, we conduct a  cross-domain evaluation from COCO to FSS-1000. The FSS-1000 dataset is originally designed for few-shot semantic segmentation, but it also supports instance-level segmentation with instance labels in 758 out of the 1,000 classes in the dataset. We select 100 classes with only instance labels from the FSS-1000 test categories (designed to be disjoint with COCO 60 base classes) as the test set. We train on COCO 60 base classes and few-shot fine-tune on FSS-1000 test classes. Since FSS-1000 has only 10 images per class, we only provide results with shot $K = \{1, 3, 5\}$. \cref{tab:fss1000} demonstrates the scalability and generalization ability of RefT to novel classes in FSS-1000.


\subsection{Ablation Study and Analysis}
\label{sec:ablation}
In this section, we present ablation studies for FSIS, gFSIS, and iFSIS. All experiments are conducted on the MS-COCO minival dataset with K=10, using ResNet-50 as the backbone. We employ standard MS-COCO metrics, specifically Average Precision (IoU=0.5 : 0.95), for both novel and base classes, denoted as nAP and bAP, respectively. Each test is executed 10 times with K examples from 10 seeds for each class, and the averaged results are reported.
\subsubsection{FSIS and gFSIS}
\noindent
\textbf{Ablation study on each component.}\quad
In Tab.~\ref{tab:abl_component}, we focus on the effectiveness of each component. For a fair comparison, we use the single-branch Mask2Former as our baseline. Incorporating our query-based second reference module results in a 1.6 improvement in novel classes. Additionally, including our image-based first reference module further enhances nAP by 1.1. It is worth noting that while longer training contributes to greater performance gains in nAP at the expense of a rapid decline in bAP, we are not sacrificing bAP for nAP. This supports our observations in Sec.~\ref{sec:pre_know}, indicating that both branches exhibit a coupled effect for novel classes.

Another interesting point is that the first reference yields a 0.9 improvement in nAP and a 0.3 improvement in bAP, while the second reference yields a 1.6 improvement in nAP but a 0.5 loss in bAP.
We assume that this may be due to the fact that the detection of base classes mainly relies on the knowledge encoded in the model parameters during base training with large amount of data, while the detection of novel classes relies mainly on the reference information provided by the support branch.
The first reference is the backbone feature aggregation. Since the two branches share the same backbone, the features of the base classes in the support branch should still be more prominent. Therefore, no loss in bAP is observed, and the additional guidance has an improvement on both base and novel classes.
The second reference is the query feature aggregation. Here, the model is encouraged to rely more on the support object query features for classification, which reduces the model bias towards the base classes. This is particularly beneficial for the classification of novel classes but may result in some loss for the base classes.

\begin{table*}[t]
\centering
\caption{Comparison with more baselines integrated with modules of previous methods on COCO with K = \{1, 3, 5, 10, 30\}. Optimal and suboptimal results are highlighted in \textbf{bold} and \underline{underline}, respectively.}
\label{tab:fair}
\setlength\tabcolsep{12pt}
\resizebox{\textwidth}{!}{%
\begin{tabular}{@{}clcccccccccc@{}}
\toprule
                               &                            & \multicolumn{10}{c}{Object Detection}                                                                                                                                                    \\ \cmidrule(l){3-12} 
                             &                            & \multicolumn{5}{c}{nAP}                                                                    & \multicolumn{5}{c}{bAP}                                                                    \\ \cmidrule(l){3-7}\cmidrule(l){8-12}
 \multirow{-3}{*}{Improve}    & \multirow{-3}{*}{Methods} & 1            & 3                & 5                            & 10             & 30               & 1           & 3                 & 5                            & 10    & 30                       \\ \midrule

 / & Mask2Former-ft-full   & 3.0 & 7.8 & 9.5    & 15.3 & 20.1 & 36.7  & \underline{36.3} & 35.9    & 36.9 & 36.2 \\  
 Backbone FA & Mask2Former w/ DCNet & 3.5 & 8.2  & 10.4     & 16.2 & 21.0 & 36.9  & 36.0   & 36.0    & 37.0   & 36.6\\ 
 Backbone FA & Mask2Former w/ FCT & 5.1 & 10.0  & 12.2    & 17.0 & 21.8 & 37.5  & 36.2 & 35.9     & 36.9 & 36.8\\ 
 Head & Mask2Former w/ Cos & 4.5 & 9.4  & 12.0    & 17.6 & 22.9   & 37.8  & 36.5 & 36.0     & 37.0 & 36.8\\ 
 Head & Mask2Former w/ Bay & \textbf{5.5} & \textbf{10.3}  & \underline{12.5}    & \underline{18.1} & \underline{23.3} & \underline{38.1}  & \textbf{36.9} & \textbf{36.2}    & \underline{37.2} & \underline{37.0}\\ 
 \cellcolor[HTML]{e9f1f6}Query FA  & \cellcolor[HTML]{e9f1f6}RefT (Ours)   & \cellcolor[HTML]{e9f1f6}\underline{5.2}         &  \cellcolor[HTML]{e9f1f6}\underline{10.2}               &\cellcolor[HTML]{e9f1f6}\textbf{13.1}                         & \cellcolor[HTML]{e9f1f6}\textbf{18.6}      &  \cellcolor[HTML]{e9f1f6}\textbf{23.4}                 & \cellcolor[HTML]{e9f1f6}\textbf{38.4}          &    \cellcolor[HTML]{e9f1f6}\underline{36.5}           & \cellcolor[HTML]{e9f1f6}\underline{36.0}                         &\cellcolor[HTML]{e9f1f6}\textbf{37.7}       &  \cellcolor[HTML]{e9f1f6}\textbf{37.2}                \\ \midrule

                             &                           & \multicolumn{10}{c}{Instance Segmentation}                                                                                                                                                    \\ \cmidrule(l){3-12} 
                             &                            & \multicolumn{5}{c}{nAP}                                                                    & \multicolumn{5}{c}{bAP}                                                                    \\ \cmidrule(l){3-7}\cmidrule(l){8-12}
\multirow{-3}{*}{Improve}    & \multirow{-3}{*}{Methods} & 1         & 3                  & 5                            & 10            & 30                & 1              & 3              & 5                            & 10             & 30               \\ \midrule
                                    
 / & Mask2Former-ft-full & 2.9 & 7.6  & 8.5    & 14.7 & 17.9 & 36.1 & 35.6  & 33.4   & 36.0 & 35.2 \\
 Backbone FA & Mask2Former w/ DCNet & 3.5 & 8.0  & 10.1     & 15.7 & 18.4  & 36.1   & 36.0 &  33.6    & 35.9  &  35.2\\ 
 Backbone FA & Mask2Former w/ FCT & 5.0 & 9.8  & 12.0    & 16.6 & 19.0 & 36.1  & 36.1 &  33.7     & 35.9 & 35.2\\ 
  Head & Mask2Former w/ Cos & 4.5 & 9.3  & 11.8    & 16.7  & 20.1 & 36.2  & 36.2 & 34.0    & 36.0  & 35.4\\ 
  Head & Mask2Former w/ Bay & \textbf{5.4} & \textbf{10.2}  & \underline{12.3}    & \underline{17.0}  & \underline{20.9}  & \textbf{36.3}   & \underline{36.4}  & \textbf{34.6}   & \textbf{36.0} & \underline{35.4}\\
 \cellcolor[HTML]{e9f1f6}Query FA & \cellcolor[HTML]{e9f1f6}RefT (Ours)   & \cellcolor[HTML]{e9f1f6}\underline{5.2} & \cellcolor[HTML]{e9f1f6}\textbf{10.2} & \cellcolor[HTML]{e9f1f6}\textbf{12.4} & \cellcolor[HTML]{e9f1f6}\textbf{17.4} & \cellcolor[HTML]{e9f1f6}\textbf{21.7} & \cellcolor[HTML]{e9f1f6}\textbf{36.3} & \cellcolor[HTML]{e9f1f6}\textbf{36.5} & \cellcolor[HTML]{e9f1f6}\underline{34.4} & \cellcolor[HTML]{e9f1f6}\textbf{36.0} & \cellcolor[HTML]{e9f1f6}\textbf{35.5} \\ \bottomrule
\end{tabular}%
}
\end{table*}

\begin{table*}[t]
\centering
\caption{Evaluation results on COCO without re-training with K = \{1, 3, 5, 10, 30\}. Optimal and suboptimal results are highlighted in \textbf{bold} and \underline{underline}, respectively.}
\setlength\tabcolsep{18pt}
\resizebox{\textwidth}{!}{%
\begin{tabular}{@{}lcccccccccc@{}}
\toprule
\multicolumn{1}{c}{} & \multicolumn{5}{c}{nAP}        & \multicolumn{5}{c}{bAP}          \\ \cmidrule(l){2-6} \cmidrule(l){7-11} 
\multicolumn{1}{c}{} & 1   & 3   & 5    & 10   & 30   & 1    & 3    & 5    & 10   & 30   \\ \midrule
\multicolumn{1}{c}{} & \multicolumn{10}{c}{Object Detetion}                              \\ \midrule
iMTFA~\cite{ganea2021incremental}                & 3.3 & -   & 6.2  & 7.1  & -    & -    & -    & -    & -    & -    \\
AirDet~\cite{airdet}               & 6.0   & 7.0 & 7.8 & 8.7 & -    & -    & -    & -    & -    & -    \\
FS-DETR~\cite{fsdetr}               & \textbf{7.0}   & \textbf{10.0}  & \underline{10.9} & \underline{11.3} & -    & -    & -    & -    & -    & -    \\
\cellcolor[HTML]{e9f1f6}RefT (Ours)          & \cellcolor[HTML]{e9f1f6}\underline{6.6} & \cellcolor[HTML]{e9f1f6}\underline{9.7} & \cellcolor[HTML]{e9f1f6}\textbf{11.3} & \cellcolor[HTML]{e9f1f6}\textbf{12.7} & \cellcolor[HTML]{e9f1f6}\textbf{13.5} & \cellcolor[HTML]{e9f1f6}\textbf{33.9} & \cellcolor[HTML]{e9f1f6}\textbf{33.7} & \cellcolor[HTML]{e9f1f6}\textbf{34.2} & \cellcolor[HTML]{e9f1f6}\textbf{33.9} & \cellcolor[HTML]{e9f1f6}\textbf{34.5} \\ \midrule
                     & \multicolumn{10}{c}{Instance Segmentation}                        \\ \midrule
iMTFA                & \underline{2.8} & -   & \underline{5.2}  & \underline{7.1}  & -    & -    & -    & -    & -    & -    \\
\cellcolor[HTML]{e9f1f6}RefT (Ours)          & \cellcolor[HTML]{e9f1f6}\textbf{6.5} & \cellcolor[HTML]{e9f1f6}\textbf{9.5} & \cellcolor[HTML]{e9f1f6}\textbf{10.8} & \cellcolor[HTML]{e9f1f6}\textbf{12.9} & \cellcolor[HTML]{e9f1f6}\textbf{13.2} & \cellcolor[HTML]{e9f1f6}\textbf{32.7} & \cellcolor[HTML]{e9f1f6}\textbf{32.5} & \cellcolor[HTML]{e9f1f6}\textbf{33.6} & \cellcolor[HTML]{e9f1f6}\textbf{32.8} & \cellcolor[HTML]{e9f1f6}\textbf{33.1} \\ \bottomrule
\end{tabular}
}
\label{tab: retrain}
\end{table*}

\begin{table}[t]
\caption{gFSOD and gFSIS results using the scaled-up backbone on COCO with K = \{1, 5, 10\}. ``-": unavailable corresponding result. }
\label{tab:scaleup}
\setlength\tabcolsep{6pt}
\resizebox{\columnwidth}{!}{%
\begin{tabular}{@{}clcccccc@{}}
\toprule 
                             &                            & \multicolumn{6}{c}{Object Detection}                                                                                                                                                    \\ \cmidrule(l){3-8} 
                             &                            & \multicolumn{3}{c}{nAP}                                                                    & \multicolumn{3}{c}{bAP}                                                                    \\ \cmidrule(l){3-5}\cmidrule(l){6-8}
\multirow{-3}{*}{Backbone}    & \multirow{-3}{*}{Methods} & 1                           & 5                            & 10                            & 1                            & 5                            & 10                           \\ \midrule 

 &   LVC~\cite{kaul2022label}   & -                           & -                            &    {18.6}                          & -                            & -                            &    {29.2}                         \\
\multirow{-2}{*}{Swin-T}  &  RefT (Ours)  &    {5.3} &    {16.8} &    {20.0}    &    {39.6} &    {37.0}   &    {37.9} \\ \midrule
 &   LVC~\cite{kaul2022label}     & -                           & -                            &    {19.0}                            & -                            & -                            &    {28.7}                         \\
\multirow{-2}{*}{Swin-S} &  RefT (Ours)   &    {5.2} &    {21.0}   &    {24.2}  &    {40.4} &    {39.8} &    {40.5} \\ \midrule

      Swin-B    &  RefT (Ours)               &  7.4                           & 20.2                           &  26.4                             &  42.6                            & 40.9                            &  41.0                            \\ \midrule 
      
                             &                            & \multicolumn{6}{c}{Instance Segmentation}                                                                                                                                                    \\ \cmidrule(l){3-8} 
                             &                            & \multicolumn{3}{c}{nAP}                                                                    & \multicolumn{3}{c}{bAP}                                                                    \\ \cmidrule(l){3-5}\cmidrule(l){6-8}
\multirow{-3}{*}{Backbone}    & \multirow{-3}{*}{Methods} & 1                           & 5                            & 10                            & 1                            & 5                            & 10                           \\ \midrule 

Swin-T          &  RefT (Ours)                   &  5.2                         &  15.4                         &  18.5                          &  37.1                         &  35.0                           &  36.2                         \\
 
Swin-S                        &  RefT (Ours)                  &  5.0                           &  19.4                         &  22.7                          &  38.3                         &  37.9                         &  38.3                         \\

Swin-B                   &  RefT (Ours)                  &  7.1                           &  20.7                            &  24.8                             &  40.2                            &  38.7                            & 39.2                            \\ \bottomrule 
\end{tabular}%
}
\end{table}

\begin{table*}[t]
\caption{Speed and Memory Analysis of Different Few-Shot Architectures.}
\resizebox{\textwidth}{!}{%
\setlength\tabcolsep{14pt}
\begin{tabular}{@{}llcccccc@{}}
\toprule  
\multirow{2}{*}{Venue} & \multirow{2}{*}{Methods} & \multicolumn{1}{l}{\multirow{2}{*}{Architecture}} & \multicolumn{1}{l}{\multirow{2}{*}{Training Strategy}} & \multicolumn{2}{c}{Train}                                   & \multicolumn{2}{c}{Inference}                                    \\ \cmidrule(l){5-6} \cmidrule(l){7-8} 
                       &                          & \multicolumn{1}{l}{}                      & \multicolumn{1}{l}{}                  & \multicolumn{1}{l}{Speed} & \multicolumn{1}{l}{VRAM} & \multicolumn{1}{l}{Speed} & \multicolumn{1}{l}{VRAM} \\ \midrule
CVPR'22                & iFS-RCNN~\cite{nguyen2022ifs}                & RPN                                       & Finetune                                    & 0.10s                             & 3543M                    & 0.14s                            & 5363M                    \\
Baseline               & Mask2Former-ft             & DETR                                      & Finetune                                    & 0.19s                            & 4973M                    & 0.10s                             & 4523M                    \\
AAAI'22  Oral          & Meta Faster R-CNN~\cite{nguyen2022ifs}         & RPN                                       & Meta                                  & 0.97s                            & 9583M                    & 0.92s                            & 8483M                    \\
Ours                   & RefT                     & DETR                                      & Meta                                  & 0.31s                            & 5655M                    & 0.11s                            & 4705M                    \\ \bottomrule  
\end{tabular}%
}
\begin{tablenotes}
        \footnotesize
        \item[*] All tested on 1 RTX 3090 with batchsize 1 and image size 1,024*1,024.  
      \end{tablenotes}
\end{table*}

\noindent
\textbf{Effect of fine-tuning CAL}\quad
To illustrate the significance of CAL in enabling DETR-like models to learn novel classes, we freeze CAL during the novel fine-tuning stage and only fine-tune the object queries and the class head. Tab.~\ref{tab:abl_cal} reveals that the model struggles to generalize with a frozen CAL, resulting in a substantial gap compared to the model that fine-tunes CAL (2.6 vs 17.4 nAP). 
Moreover, the findings suggest that additional learnable parameters are unnecessary to ensure adequate transferability to the novel domain, as fine-tuning an existing CAL can achieve both novel class generalization and base knowledge retention.

\noindent
\textbf{Support features in first reference}\quad
In Tab.~\ref{tab:abl_feat_sel}, we conduct an ablation study on various features used to compute adaptive class prototypes, including three stages of features from ResNet-50 (denoted as Res3, Res4, and Res5) and the flattened multi-scale features from the first transformer encoder layer (denoted as Enc1). 
Utilizing support features of a larger scale results in a notable performance improvement (17.2 vs 16.4 nAP), potentially due to the increased detection of smaller objects. 
Employing multi-scale features further contributes to a 0.2 nAP gain.

\noindent
\textbf{Support features in second reference}\quad
To showcase the efficacy of our query-based aggregation in the second reference, we implement an image-based version that employs the same adaptive class prototypes from the first reference as support guidance and uses cross-attention as the aggregation operation. Tab.~\ref{tab:abl_query_or_img} demonstrates that the query-level enhancement module outperforms the feature-level one, resulting in a 1.1 nAP increase. This validates the superiority of our RefT framework over a purely feature-level enhancement framework.

\noindent
\textbf{Pooling in first reference}
In Tab.~\ref{tab:abl_pooling}, we compare various pooling methods for extracting pertinent information from support images. As the accuracy of the feature region improves, the performance increases correspondingly. This is attributed to the more accurate instance features without noise, which provide more distinctive class prototypes for the query branch classification.

\noindent
\textbf{Query selection in second reference}
In Tab.~\ref{tab:abl_query_sel}, we compare the results of selecting top 10 object queries in second reference sorted by scores or mask IoU. The results are comparable when k is small, as object queries exhibit higher accuracy for both classification and segmentation.

\noindent
\textbf{Number of Object Queries in Second Reference} In the second reference, we choose the top-k object queries for each support image. Tab.~\ref{tab:abl_query_num} compares the performance impact of different k values. The results indicate that when $k \leq 50$, a relatively stable outcome close to the best can be achieved. This finding aligns with Tab.~\ref{tab:iou_table}, as the support branch mask quality is ensured when $k \leq 50$.

\noindent
\textbf{Comparison with previous feature-level aggregation module.} Meta-DETR~\cite{zhang2022meta} introduces a Correlational Aggregation Module (CAM) for concurrent aggregation of query features and support class prototypes, akin to our first reference module. The differences are twofold: (1) the support set organization varies. Meta-DETR divides all classes into class sets, requiring multiple forward passes, while the first reference handles all classes simultaneously. (2) The classification loss differs. Meta-DETR uses a sigmoid binary cross-entropy loss, transforming the classification task into matching between query and support classes, whereas we utilize the standard softmax cross-entropy loss. As demonstrated in Tab.~\ref{tab:cam}, replacing our first reference module with CAM and maintaining the support set and loss in line with Meta-DETR leads to a substantial 9.2 bAP decrease (36.0 vs. 26.8). Even with our support set and loss function unchanged, bAP drops by 3.7 (36.0 vs. 32.3). We speculate that CAM primarily handles class matching in Meta-DETR, while the class head operates in a class-agnostic manner. Consequently, the base class feature representations are not stored in the class head as in prior works, resulting in suboptimal bAP outcomes.

\noindent
\textbf{Effect of fine-tuning different layers.}
In the main results, we only unfreeze CAL, object queries, and the class head during novel fine-tuning to enable the model to learn novel classes without forgetting previous knowledge. Tab.~\ref{tab:finetune} presents a detailed investigation of how fine-tuning different layers impacts the results. We observe that when CAL is frozen, even if both the pixel decoder and transformer decoder are fine-tuned, the model struggles to learn novel knowledge (2.5 nAP). Moreover, fine-tuning more parameters on extremely limited samples can cause overfitting, leading to a significant decline in bAP (-18.8$\downarrow$). Although fine-tuning layers in the feature extractor allows the model to generalize,  fine-tuning CAL with fewer parameters achieves better results for both novel (+5.7$\uparrow$) and base classes (+3.1$\uparrow$).

\subsubsection{iFSIS}
\noindent
\textbf{Ablation of the Class-Enhanced BKD Loss.}
In Tab.~\ref{tab:abl_ifsis}, we first use L2 regularization to constrain the parameters of CAL as our baseline. Using BKD Loss~\cite{dong2022incremental} recovers bAP by 3.8 with a drop in nAP (-1.4$\downarrow$) while applying our Class-Enhanced BKD Loss significantly recovers bAP (+5.8$\uparrow$) with a marginal drop in nAP (-0.4$\downarrow$).

\noindent
\textbf{Employing different class heads.}
 In Tab. \ref{tab:classhead}, we also experiment with the logit classifier based on Bayesian probabilities proposed by the recent state-of-the-art iFS-RCNN~\cite{nguyen2022ifs}. Although performance improvements are observed in both nAP(+1.6) and bAP(+0.8), they are not as significant as in region-based methods. This suggests that there is potential for further improvement in DETR-like~\cite{detr} models for iFSIS.

\subsection{Fair Comparison with Previous Methods}
\label{sec:fair}
To ensure a fair comparison, we transfer the modules proposed in the RPN-based methods to the query-based architecture, \ie, Mask2Former. The transferable methods mainly include those that improve the backbone feature aggregation and the predictor head. The methods that improve RPN and ROI-feature aggregation cannot be transferred to the query-based architecture. Specifically, we transfer the module from DCNet (\ie, Dense Relation Distillation Module) and FCT (\ie, Fully Cross Transformer) for improving backbone feature aggregation, and iMTFA (\ie, Cosine-Similarity Head) and iFS-RCNN (\ie, Bayesian Head) for improving the classification head. \cref{tab:fair} shows that the improvement brought by these methods is slightly lower than or comparable to our approach. However, it is worth noting that our method is not in conflict with these methods. With the addition of these heads, our method can achieve even higher results. Our goal is to provide a simple query-based baseline for (general/ incremental) FSIS to avoid the overfitting issue of RPN. Therefore, we propose query-level feature aggregation as a replacement for ROI feature aggregation and try to solve the challenge of query-based architecture in incremental setting. We hope that this will encourage further exploration of query-based structures in future research.

\subsection{Generalizability without Re-training}
\label{sec:retrain}
RefT employs a vanilla class-specific classification head, which means it does not support the "without re-training" setting. However, by simply adding a learnable support class embedding for each support class on the support branch query features during training, RefT can easily adapt to the "without re-training" setting. By doing this, the classification head no longer predicts the score of a fixed category, but instead predicts the probability of whether the object belongs to the given support class. The model dynamically classifies based on the input support classes. Is is worth noting that our second reference module is functionally almost equivalent to FS-DETR~\cite{fsdetr} when adapted to "without re-training" setting. The only difference is that we implement cross-attention to input support queries and class embeddings, while FS-DETR concatenates them with object queries. 

Since FS-DETR has not been open-sourced, in order to provide a comparative result in the "without re-training" setting, we did not evaluate on new dataset but still on the COCO dataset. As demonstrated in \cref{tab: retrain}, the class-agnostic training strategy greatly enhances the model generalization ability for novel classes especially in low-shot situations, but at the cost of a decrease in base class performance and a gradual reduction in performance improvement as the number of shots increases. FS-DETR also exhibits similar behavior according to its reported results.

\subsection{Scale up}
\label{sec:scale_up}
In Tab.~\ref{tab:scaleup}, we further showcase the effectiveness and generalization capability of RefT by scaling up the backbone. Using more powerful Transformer backbone models, such as Swin-T, Swin-S, and Swin-B, our approach continues to generalize well and avoids severe overfitting.

\begin{figure}[h]
\centering
\includegraphics[width=0.5\textwidth]{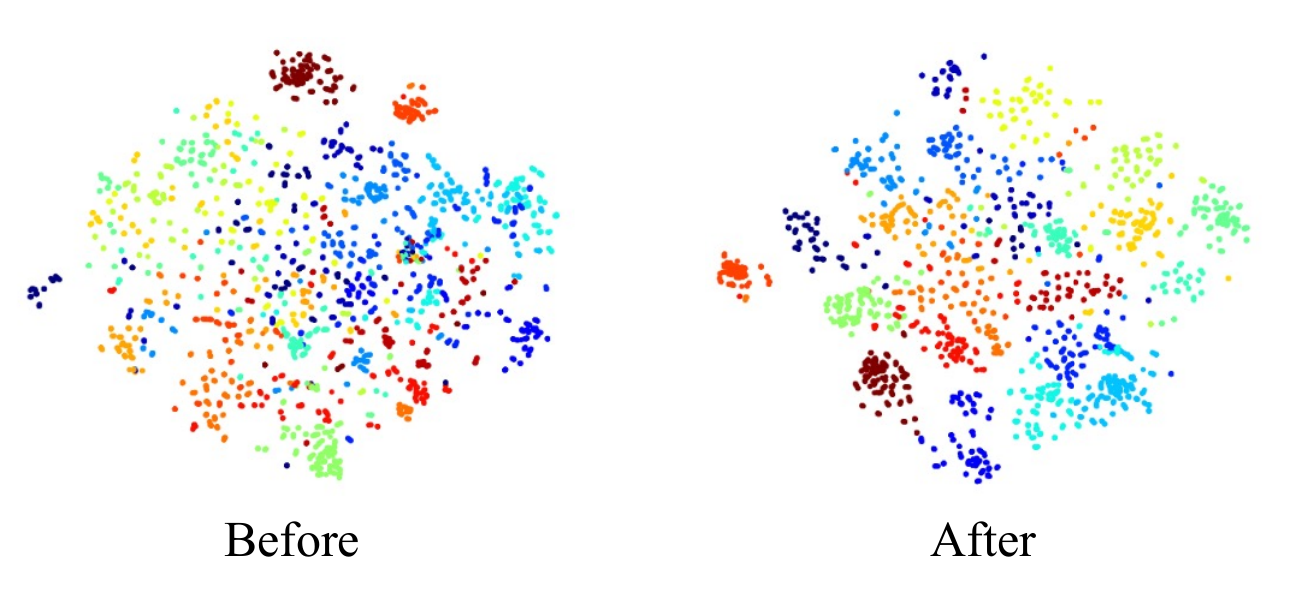}
\caption{\textbf{t-SNE visualization of object queries belonging to COCO 20 novel classes.} The results are obtained from the support branch before and after novel fine-tuning.
} \label{fig:tsne}
\end{figure}

\begin{figure}[h]
\centering
\includegraphics[width=0.5\textwidth]{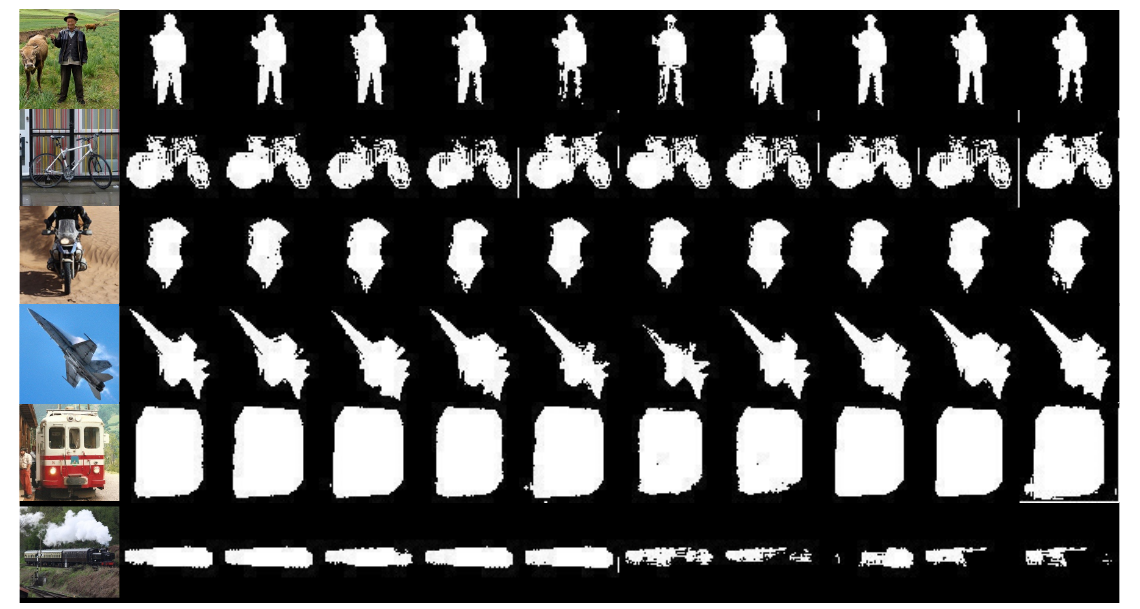}
\caption{\textbf{Visualization of the mask head predictions of support branch object queries.} Object queries that achieve top-10 IoU with the ground truth masks are shown.} \label{fig:loc}
\end{figure}

\begin{figure}[h]
\centering
\includegraphics[width=0.5\textwidth]{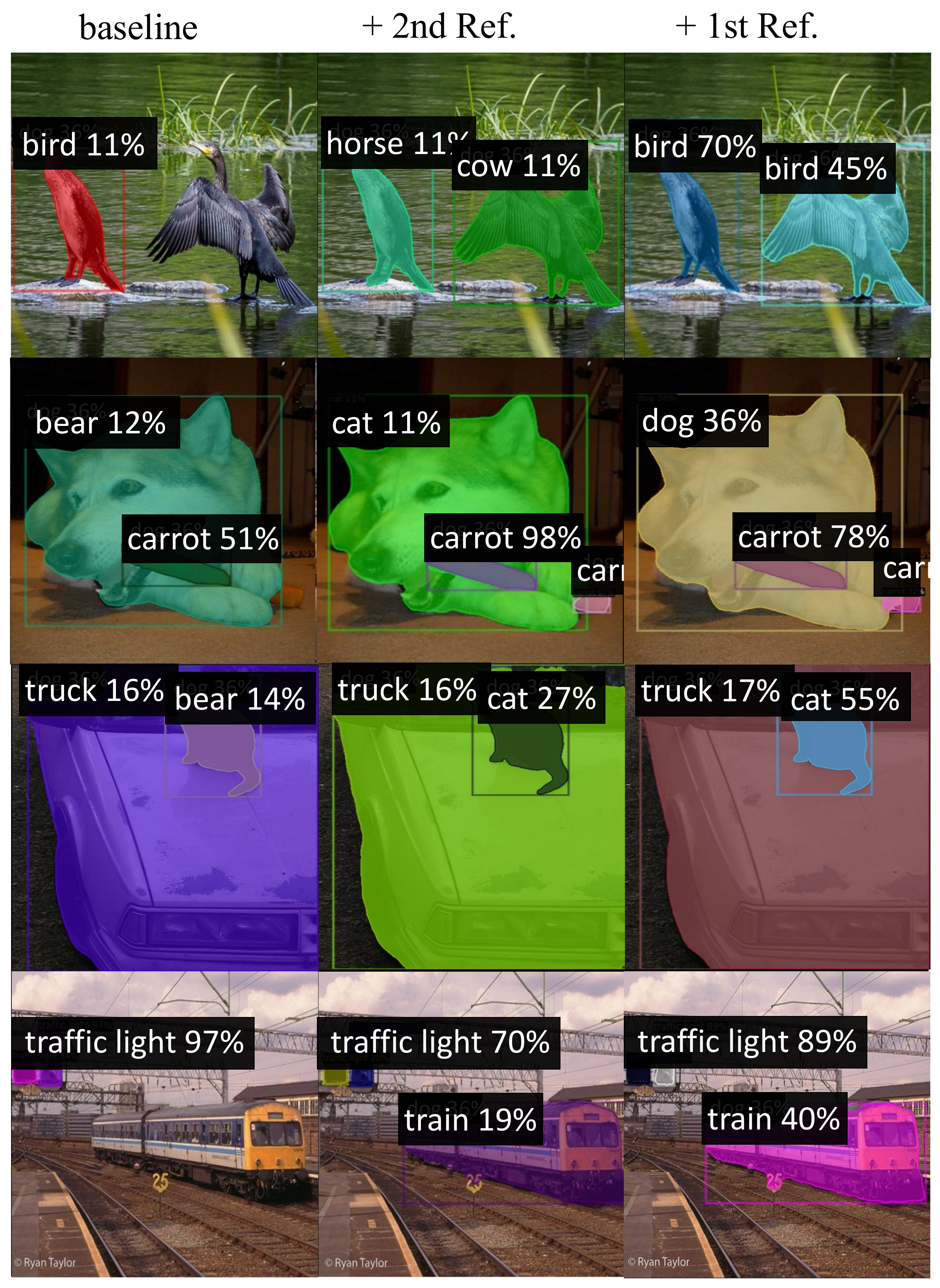}
\caption{\textbf{Visualization of 10-shot FSIS results on COCO minival set.} Results of novel classes are mainly displayed. } \label{img}
\end{figure}

\subsection{Visualization and More Analysis}
\noindent
\textbf{Understanding the effect of second reference}\quad
In Fig.~\ref{fig:tsne}, we visualize the t-SNE results of object queries for COCO's 20 novel classes. We observe that the object queries of novel classes can be roughly distinguished even without the novel fine-tuning process, and the clustering becomes more apparent after fine-tuning. 
In our second reference module, we utilize this \emph{support query categorization} to guide the query branch classification. In Fig.~\ref{fig:loc}, we visualize the mask head predictions of support branch object queries from the model before the novel fine-tuning stage. Only object queries with top-10 IoU values compared to ground truth masks are shown. We find that the object queries of novel classes encode accurate localization information even without the novel fine-tuning process. We leverage this \emph{support query localization} in our second reference module to provide guidance for the query branch localization.

Visualizations for all three settings are presented. Predictions for base class instances are shown in Fig.~\ref{fig:baseimg}, while those for novel class instances are displayed in Fig.~\ref{fig:novelimg}. 
Although class misclassification remains prevalent, most predicted masks are accurate across all settings. 
Even in the most challenging iFSIS setting, numerous novel instances are recalled.

\noindent
\textbf{Additional Qualitative Results.}\quad In Fig.~\ref{img}, we provide visual results on MS-COCO datasets corresponding to Tab.~\ref{tab:abl_component}. We observe that incorporating our query-based second reference effectively reduces both missed and misclassified instances, primarily due to the guidance from \emph{support query localization} and \emph{support query categorization}. By aligning query features with adaptive class prototypes, the inclusion of our first reference further decreases misclassification between highly correlated classes.

\subsection{Computational Cost}

\noindent
\textbf{Parameter and GFLOPs.}\quad Compared to the robust Mask2Former baseline, RefT achieves a notable +1.7$\uparrow$ increase in nAP with only a 5.8\% increment in GFLOPs (226.2 \vs 239.4) and a 6.2\% rise in parameters (43.7 \vs 46.4), while processing a 1,024 × 1,024 image input.

\noindent
\textbf{Speed.}\quad During training, RefT is slightly slower than the fine-tuning-based iFS-RCNN (0.10 \vs 0.31) due to the additional forward pass required to process a randomly sampled support set, which guides the query branch. However, RefT outperforms the meta-based Meta-FRCN (0.97 \vs 0.31) by using object queries, thereby eliminating the need for pixel space processing.

\noindent
\textbf{Inference Efficiency.}\quad During inference, RefT becomes more efficient as the support set is preprocessed and fixed.

\noindent
\textbf{VRAM.}\quad The memory cost of RefT is limited due to the following factors: \textbf{\textit{1)}} ReFT mainly finetunes class-specific parameters while keeping most parameters fixed; \textbf{\textit{2)}} RefT adopts a smaller 320 × 320 image size in the support branch, significantly reducing memory consumption compared to the query branch.

\section{Conclusion} \label{section:con}
In this paper, we present a simple and unified baseline for few-shot instance
segmentation, namely \emph{Reference Twice} (RefT). We carefully examine the mask-based DETR framework in FSIS and identify two key factors named \emph{support query localization} and \emph{support query categorization}. 
Specifically, we first design a mask-based dynamic weighting module to aggregate support features and then propose to link object queries for better calibration via cross-attention. 
Additionally, RefT can be easily extended to all three settings including FSIS, gFSIS, and iFSIS with a simple Class-Enhanced BKD loss to solve the difficulty of adapting DETR-like models to incremental settings without selective search or logits distillation. To the best of our knowledge, we are the first unified architecture to support three different few-shot settings
Despite its simplicity, RefT achieves state-of-the-art or second-best performance on MS-COCO and LVIS benchmarks, across all settings and all shots. 
We hope our framework can be a solid baseline for instance-level few-shot segmentation problems.

\noindent
\textbf{Limitation and Future Work.}
Currently, RefT fails to perform well in the one-shot setting. One shot makes the guidance of the reference-twice modules unreliable because the support features deviate from the true class distribution. One potential solution is to improve the single object matching ability to fix this issue.



\section{acknowledgement} \label{section:ack}
This work was supported by the Zhejiang Provincial Science and Technology Planning Project (No. 2024C01172).

\begin{figure*}[h!]
\centering
\includegraphics[width=\textwidth]{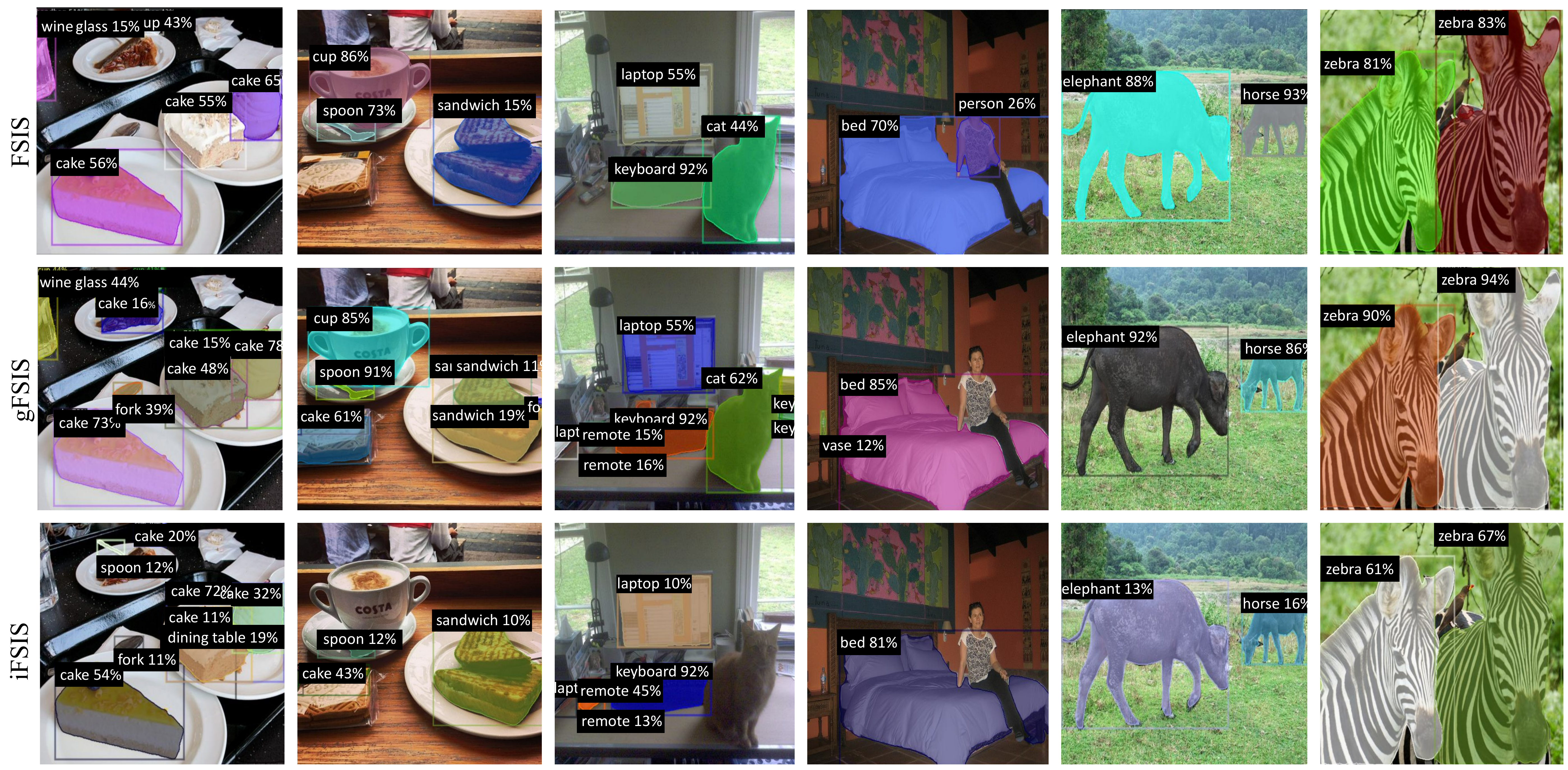}
\vspace{-3mm}
            \caption{\textbf{Visualization of results for base classes on COCO with K=10.}} \label{fig:baseimg}
\end{figure*}

\begin{figure*}[h!]
\centering
\includegraphics[width=\textwidth]{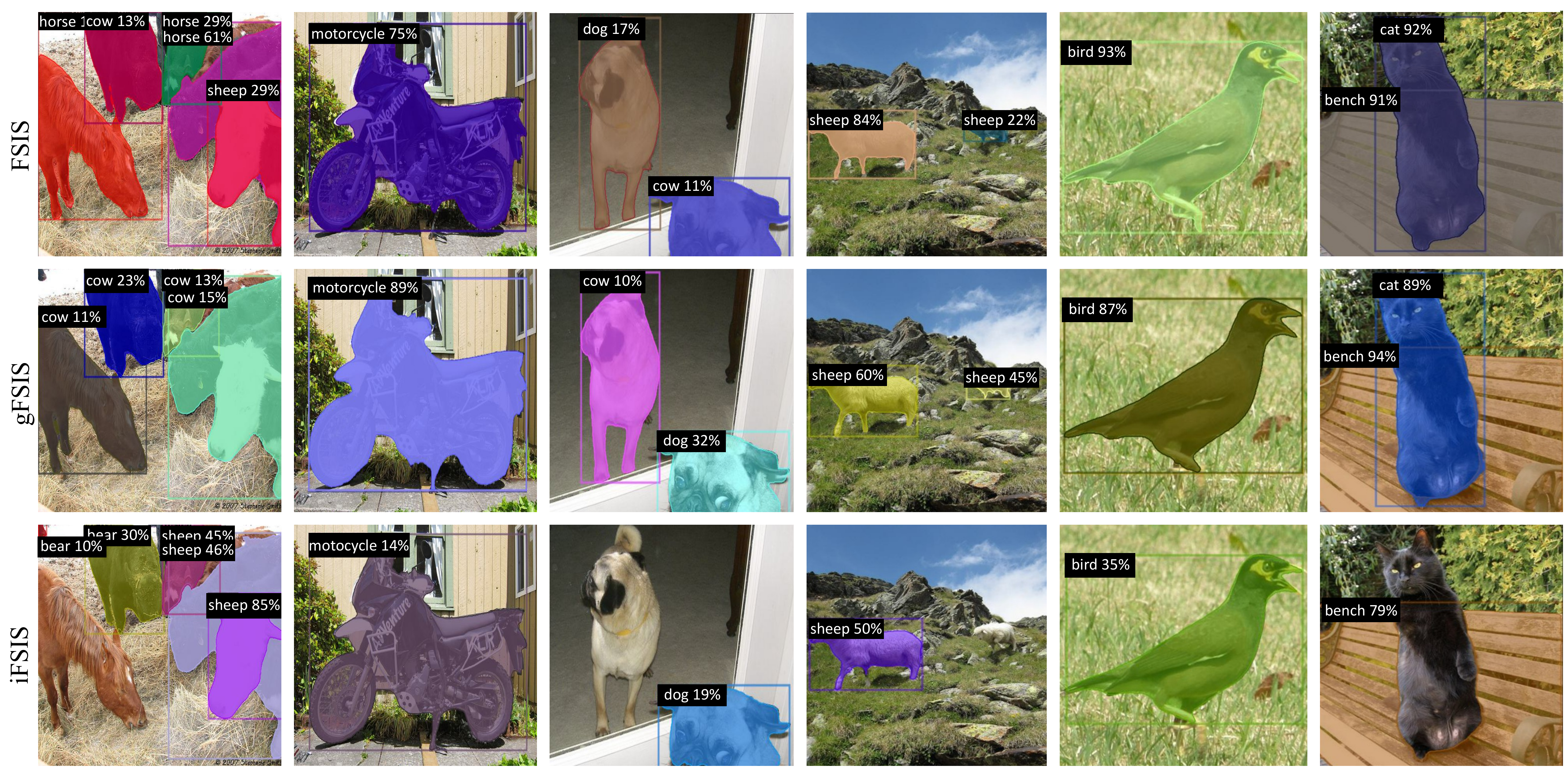}
\vspace{-3mm}
\caption{\textbf{Visualization of results for novel classes on COCO with K=10.}} \label{fig:novelimg}
\end{figure*}
\ifCLASSOPTIONcaptionsoff
  \newpage
\fi


\newpage 

{
\bibliographystyle{IEEEtran}
\bibliography{IEEEabrv,main}
}

\end{document}